\documentclass[preprint,3p,times]{elsarticle}

\usepackage{amssymb}
\usepackage{amsmath}
\usepackage{url}
\usepackage{subcaption}
\usepackage{xcolor}

\journal{Computational Material Science}

\begin{document}

\begin{frontmatter}



\title{High-Throughput Unsupervised Profiling of the Morphology of 316L Powder Particles for Use in Additive Manufacturing} %% Article title

\author[1]{Emmanuel Akeweje}
\author[2]{Conall Kirk}
\author[2]{Chi-Wai Chan}
\author[3]{Denis Dowling}
\author[1]{Mimi Zhang}

\affiliation[1]{organization={School of Computer Science and Statistics, Trinity College Dublin}, country={Ireland}}
\affiliation[2]{organization={School of Mechanical and Aerospace Engineering, Queen's University Belfast}, country={UK}}
\affiliation[3]{organization={School of Mechanical and Materials Engineering, University College Dublin}, country={Ireland}}

%% Abstract
\begin{abstract}
Selective Laser Melting (SLM) is a powder-bed additive manufacturing technique whose part quality depends critically on feedstock morphology. However, conventional powder characterization methods are low‐throughput and qualitative, failing to capture the heterogeneity of industrial‐scale batches. We present an automated, machine learning framework that couples high‐throughput imaging with shape extraction and clustering to profile metallic powder morphology at scale. We evaluate three complementary approaches. The first uses an orientation-invariant variational autoencoder (VAE) to learn shape features that are insensitive to particle rotation, after which the resulting latent representations are grouped using clustering methods. The second applies classic shape descriptors -- centroid distance functions, Fourier descriptors, and Zernike moments -- followed by k-means or Gaussian mixture clustering to group particles with similar outlines. The third treats each particle’s radial profile as a continuous function and clusters these functional representations using the GPmix algorithm, allowing subtle differences in lobing and eccentricity to be captured. Across a dataset of approximately 126,000 powder images (0.5 -- 102 $\mu$m diameter), internal validity metrics identify the Fourier-descriptor + k-means pipeline as the most effective, achieving the lowest Davies–Bouldin index and highest Calinski–Harabasz score while maintaining sub-millisecond runtime per particle on a standard desktop workstation. The VAE provides orientation-invariant representations but at significantly higher computational cost, whereas the functional-data approach captures finer shape variations but requires a more complex configuration. Although the present work focuses on establishing the morphological-clustering framework, the resulting shape groups form a basis for future studies examining their relationship to flowability, packing density, and SLM part quality.
Overall, this unsupervised learning framework enables rapid, automated assessment of powder morphology and supports tracking of shape evolution across reuse cycles, offering a path toward real-time feedstock monitoring in SLM workflows.
\end{abstract}

% % \includegraphics{grabs}

% \begin{highlights}
% \item Automated machine learning framework for large-scale Selective Laser Melting (SLM) powder morphology profiling
% \item Evaluated three unsupervised pipelines: using variational autoencoder, shape descriptors, and functional clustering
% \item Fourier-descriptor + k-means achieved best accuracy, speed, and efficiency
% \item Framework distinguishes spherical, lobed, elongated, and defective particles
% \item Enables real-time monitoring of powder quality across reuse cycles in SLM
% \end{highlights}

%% Keywords

\end{frontmatter}


\section{Introduction}
Selective Laser Melting (SLM), a form of Powder Bed Fusion (PBF), has emerged as a transformative additive manufacturing (AM) technique for producing net-shape metallic components with complex geometries that are otherwise unachievable through traditional subtractive or forming processes. Its adoption spans critical industries, including aerospace \citep{seabra_selective_2016}, nuclear \citep{jialin_selective_2017}, and biomedical \citep{kelly_fatigue_2019}, due to its ability to fabricate intricate parts in stainless steels, titanium alloys, shape-memory alloys, and magnesium alloys \citep{kirk_microstructure_2025, xie_enhancing_2024, wang_preparation_2025, shi_effects_2024}. The SLM process is governed by intricate interactions between a high-energy laser and a layer of metal powder, where over one hundred parameters affect the final part quality, ranging from laser power to scan speed.

Powder–laser interaction plays a central role in this process: particle size, shape, and packing density influence local absorptivity, energy uptake, and melt-pool initiation, which in turn affect the stability of the track and the likelihood of defects such as balling, lack of fusion, or keyholing. While these early-stage interactions have been discussed extensively in reviews of SLM melt-pool physics \citep{gunasekaran2022brief, sefene2022state}, the influence of powder morphology itself has received comparatively limited attention. Variations in particle shape, size distribution, surface roughness, and the effects of powder recycling can alter powder-bed density, porosity, and flowability, ultimately affecting layer uniformity and the mechanical performance of printed components \citep{tan_overview_2017, avrampos_review_2022}. Given that morphology directly contributes to both how the powder absorbs laser energy and how it packs within the bed, a more quantitative understanding of powder shape distributions is needed to reduce variability and improve build reliability in SLM.

Traditional powder characterization workflows -- most commonly optical microscopy or scanning electron microscopy (SEM) coupled with manual or semi-automated measurements in tools such as ImageJ -- tend to emphasize qualitative sphericity inspection and coarse size metrics \citep{dilip_influence_2017, qu_anisotropic_2022}. These approaches are insufficient to capture the high-dimensional complexity of powder morphology at industrially relevant scales. Manual inspection of a small subset cannot represent batch heterogeneity (for reference, $\sim$1\,kg of 30\,$\mu$m 316L powder contains $\approx 9 \times 10^{9}$ particles, assuming $\rho \approx 8$\,g\,cm$^{-3}$ and monodisperse 30\,$\mu$m spheres). Moreover, analysis based on a limited number of samples can introduce subjective interpretation, potentially overlooking important details such as the diverse types of powder irregularities. This gap underscores the need for an automated, scalable technique that quantitatively analyzes full shape distributions and identifies salient morphological subgroups before printing.

In this work, we introduce an unsupervised learning framework that combines high-throughput imaging with advanced morphology-analysis methods to classify two-dimensional binary images of individual powder particles at scale. We cluster particles based on features that are invariant to translation, rotation, and scale, including centroid distance functions, Fourier and Zernike moments, and steerable convolutional autoencoder embeddings. The proposed approach is aimed at providing operators with a rapid assessment of particle morphology, thereby enabling predictive insights into powder behavior such as flowability and packing density. While powder characteristics represent only one of several factors influencing build outcomes, improved powder quality assessment can contribute to reducing variability in the powder bed and, in turn, support efforts to minimize scrap and mitigate the likelihood of failed builds. Additionally, it enables tracking of powder morphology across successive recycling cycles, improving material efficiency and reliability in SLM workflows.

The remainder of the paper is organized as follows: Section \ref{background} reviews prior work on powder feedstock characterization and powder shape-analysis techniques; Section \ref{morphology} describes our methodology, covering image preprocessing, the computation of invariant shape descriptors, and the three clustering pipelines; Section \ref{Results} presents experimental results on 316L powder particles, evaluating cluster quality for the three different clustering methods; and Section \ref{Conclusion} concludes with a discussion of our findings, practical implications for real‐time SLM feedstock control, and directions for future research.

\section{Related Work}\label{background}
%\subsection{Powder Production and Morphology}
Powder feedstock for AM is typically produced via atomization methods -- water, gas, or plasma -- which yield particles with distinct shape characteristics and satellite content. Water atomization often generates highly irregular particles, gas atomization produces moderately spherical powders with fewer satellites, and plasma atomization yields the most uniform and spherical particles \citep{li_densification_2010, dawes_introduction_2015}. Recent innovations in ultrasonic vibration machining demonstrate the production of uniform, micron-sized metal chips with tight size tolerances and high sphericity, leading to improved microhardness in printed A356 parts \citep{WANG2024103993}. Additionally, ultrasonic atomization of scrap maraging-steel can regenerate satellite-free, spherical powders comparable to commercial feedstocks \citep{Lagoda2025}, while plasma spheroidization models have been developed to predict morphology transformations in refractory powders such as tungsten \citep{Sharma2025}.

%\subsection{ Methods for Powder Analysis}
Traditional morphology metrics -- mainly qualitative assessments of sphericity via SEM -- are limited by low throughput, typically capturing only hundreds to a few thousand particles per dataset and therefore introducing sample bias \citep{dilip_influence_2017, qu_anisotropic_2022}. Although SEM remains widely used, the data acquisition process is inherently slow and not well suited to large powder batches. Synchrotron X-ray computed tomography has been applied to aluminum-alloy powders to obtain three-dimensional morphology and internal porosity information, providing true sphericity and size distribution metrics at scale \citep{Ouyang2023}. However, X-ray CT requires extensive sample preparation, long imaging and reconstruction times, and yields relatively few analyzable particles per scan, making it impractical for routine or in-process morphology assessment \citep{xue20253d}. Laser diffraction, while efficient for particle size analysis, cannot evaluate morphology at all. Statistical models combining particle-size distributions with elliptic Fourier series have shown that shape metrics such as aspect ratio and circularity influence flowability across alloys ranging from AlSi10Mg to Inconel 718 \citep{Murtaza202410853}, but their effectiveness still depends on adequate sampling. These constraints underscore the need for high-throughput acquisition methods. The Malvern Morphologi 4 particle analyzer provides a practical alternative, generating segmented 2D binary images of tens of thousands of particles in under 30 minutes with minimal sample preparation; its automated workflow also ensures consistency between batches and can be operated with basic training.

%\subsection{ Advanced Morphology Characterization}
Recent work has increasingly applied modern computer-vision methods to the characterization of powder feedstocks. A deep learning model based on generative adversarial network architecture \citep{Krishna2024} has been used to segment individual particles in microscopy images and automatically extract size and basic shape metrics such as circularity and aspect ratio, offering a faster alternative to conventional laser-diffraction measurements. Price et al. \citep{price2025dualsight} further developed a two-stage segmentation framework that produces cleaner particle outlines in SEM images, enabling more reliable morphology measurements. Deep-learning approaches have also been extended to in-situ LPBF monitoring, where models can automatically identify defects associated with recoating and powder-bed irregularities \citep{korzeniowski2024development}. Complementing these methods, classical shape descriptors based on two-dimensional moment invariants have been shown to distinguish powders with subtle differences in particle form beyond simple size metrics \citep{harrison2019use}. Keypoint-based image descriptors have likewise been demonstrated for classifying powder materials by capturing both particle morphology and surface texture, performing comparably to segmentation-based particle-size analysis \citep{decost2017characterizing}.

%\subsection{ Influence of Morphology on Powder Behavior}
Particle morphology strongly influences the bulk rheology of metal powders. Spherical particles typically exhibit good flowability and high packing density because of reduced inter-particle friction, whereas fused or agglomerated particles disrupt packing, increase void space, and hinder flow. Elongated particles can interlock and promote anisotropic packing, while the presence of satellites increases surface roughness and can reduce both flowability and spreadability \citep{radchenko2022requirements}. These morphological classes therefore play a critical role in determining the overall performance of SLM powders, underscoring the need for detailed morphology profiling. Brika et al. \citep{brika_influence_2020} found that increased particle sphericity improved flowability and packing density, leading to higher printed part density. Similarly, Mehrabi et al. \citep{mehrabi2023investigation} compared gas-atomized and hydride–dehydride powders and showed that the gas-atomized powder, with its more spherical morphology, exhibited higher flowability across multiple metrics (angle of repose, Hausner ratio). High‑speed imaging and microtomography studies show that spherical particles exhibit greater radial drag ahead of the melt pool, affecting denudation width and layer stability \citep{Fedina2022720}. Discrete element method simulations confirm that non‑spherical particles reduce packing density and elevate shear stresses during recoating \citep{Yim2023}. In binder‑jet sintering of WC–Co, cubic versus spherical particle shapes yield markedly different capillary forces and densification kinetics \citep{Chaithanya_Kumar2024}. Within SLM, increased powder bed density -- achieved with highly spherical Ti‑6Al‑4V powders -- correlates with reduced void content, enhanced surface finish, and improved dimensional accuracy \citep{brika_influence_2020, soltani-tehrani_effects_2022}.

%\subsection{ Indicating What Constitues Good Clustering}
In powder-based AM processes, morphological uniformity is widely regarded as a key indicator of powder quality, with the ideal case being a high proportion of regular spherical particles and a minimal fraction of irregular shapes. The clustering framework developed here quantifies the fraction of particles assigned to each morphological class, providing a physically interpretable measure of uniformity. Clusters corresponding to non-ideal particle shapes highlight morphological defects known to reduce flowability, packing density, and spreading performance, whereas clusters of regular spherical particles represent the desired condition. The 316L powder analyzed in this study has already been used in our previous work to produce high-quality SLM parts \citep{kirk_microstructure_2025}, indicating that its quality is sufficient for printing. Although we do not include additional external validation tests in this methodological study, the combination of cluster-based morphology profiling and prior successful printing offers practical confidence in the relevance and interpretability of the proposed framework.

%\subsection{ Effects of Powder Recycling}
Economical SLM workflows commonly depend on powder reuse, with material repeatedly cycled across successive builds for sustainability and cost efficiency \citep{cordova2019revealing}. However, repeated recycling has been shown to alter powder morphology, size distribution, composition, and flowability, ultimately impairing additive manufacturing performance. Studies of Ti-6Al-4V powders report decreased satellite content but increased core deformation and oxygen uptake after multiple LPBF cycles, producing higher strength but reduced ductility \citep{Zhuo2025}. Similarly, ultrasonic-plasma atomization of NiTi at higher vibration frequencies increases surface oxidation while maintaining ASTM F3049 size and shape standards \citep{Sojoodi2025}. Recycling of 316L stainless steel in both Electron Beam Melting and LPBF processes leads to a rise in feedstock defects and a corresponding decline in tensile and fatigue properties \citep{popov_effect_2018, delacroix_influence_2022, sutton_evolution_2020}.

In industrial settings, waste powder collected after each build is typically sieved to remove spatter, agglomerates, and oversized or undersized particles, thereby restoring the particle size distribution to an acceptable range for reuse. However, although sieving can control particle size, there is no equivalent process for restoring particle morphology. Irregular, elongated, fused, or otherwise degraded particles tend to accumulate over successive cycles because removing them selectively would require discarding large volumes of usable powder. This absence of a morphology-conditioning step highlights the need for quantitative morphology analysis methods that can detect when the shape population begins to deviate from predominantly well-formed particles. 

%\subsection{ Gaps and Motivation}
Despite established links between powder morphology and part performance -- including effects on density, mechanical properties, and fatigue behavior -- existing characterization methods lack the throughput and scalability necessary for industrial deployment \citep{brika2020influence, cordova2019revealing}. Manual or low‑volume assessments cannot capture the heterogeneous nature of large powder batches, and while machine‑learning techniques have been applied to SEM images for manufacturer classification \citep{decost_computer_2017} and supervised 3D shape classification \citep{zhou_intelligent_2022}, they do not provide an automated, unsupervised solution for 2D morphology profiling at scale. This gap motivates our framework, which leverages high‑throughput imaging and invariant feature extraction for the assessment of the morphological subgroup distribution of powder particles, prior to the use of the powder batch in printing.

\section{Methodology} \label{morphology}
\subsection{Powder Particle Image Acquisition and Dataset Preparation}
The 316L stainless steel powder investigated in this study was produced by gas atomization, a process known to generate predominantly spherical particles with a narrow size distribution. Approximately 7mm$^3$ of powder was randomly dispersed onto the sample holder to ensure unbiased sampling. Imaging was performed using a Malvern Morphologi 4 particle analyzer equipped with a 20× objective lens. The instrument's proprietary \textit{fsSharpEdge} mode performs on-device edge enhancement and illumination normalization, producing uniformly contrasted images with sharply defined particle boundaries. Because Morphologi 4 employs an adaptive imaging protocol, the effective pixel size is not fixed; across the dataset, pixel sizes span approximately 0.008--0.089$\mu$m/px (5th--95th percentile). This acquisition approach reliably yields well-isolated particle silhouettes, resulting in 125,874 high-quality 2D grayscale images for subsequent morphological analysis.

Although these images are 2D projections of inherently 3D particles, the large number of particles provides statistically meaningful insight into the prevalence of different particle morphologies. These images allow robust quantification of morphological uniformity and the distribution of ideal versus non-ideal shapes that affect powder flow, spreading behavior, and layer quality in AM processes.

\subsection{Image Preprocessing}
\begin{figure}[!h]
    \centering
    \includegraphics[width=0.8\linewidth]{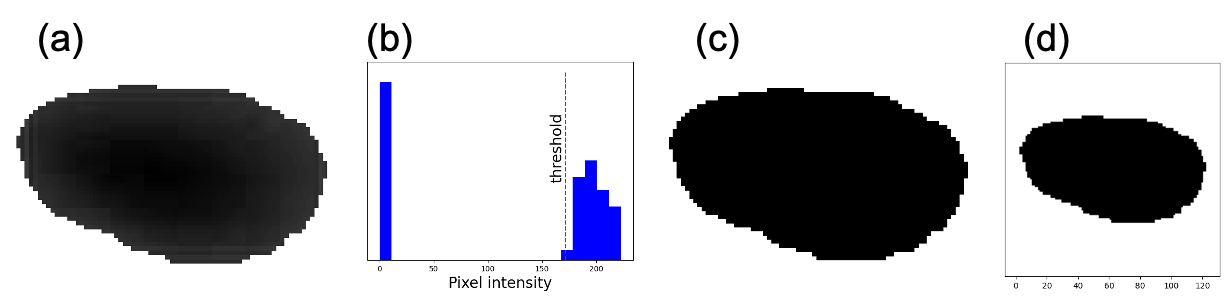}
    \caption{(a) Original grayscale micrograph of a single particle captured on the Malvern Morphologi 4. (b) Intensity histogram with Otsu's threshold (vertical dashed line) used to separate foreground particles from background. (c) Binary mask generated by applying the threshold, isolating the particle silhouette. (d) Final normalized input: the mask is centered, padded, and resized to a 128 × 128-pixel canvas, ensuring consistent spatial dimensions for VAE.}
    \label{fig:img-processing}
\end{figure}

Given the consistent contrast and clean silhouettes produced by the imaging system, image preprocessing begins by converting each grayscale particle image to a binary form so that the particle silhouette is cleanly separated from the background; see Figure \ref{fig:img-processing}. First, Otsu’s thresholding algorithm examines the grayscale histogram and selects a cutoff value that maximizes the variance between the dark (foreground) and light (background) pixel classes. Every pixel darker than this automatically determined threshold becomes part of the binary mask (particle), while brighter pixels become background. Because the \textit{fsSharpEdge} acquisition mode already yields well-isolated particles with uniform contrast, no additional background correction, watershed separation, or morphological smoothing is required. Finally, each binary mask is resized to a fixed width and height only for use in the VAE, which requires inputs with consistent spatial dimensions. The classical shape descriptors -- centroid distance, Fourier, and Zernike moments -- are computed directly from the original binary masks without resizing, since these descriptors are intrinsically invariant to scale as well as rotation and translation.

\subsection{Shape Descriptors}\label{shape-descriptors}
Feature extraction converts each binary particle image into a compact, numerical signature that is invariant to translation, rotation and scale -- key requirements for comparing arbitrary powder shapes. We employ three complementary descriptors: the Centroid Distance Function (CDF), Fourier Descriptors (FDs) and Zernike Moments (ZMs), each capturing distinct aspects of boundary geometry, frequency content and global form.

\begin{figure}[!ht]
     \centering
     \includegraphics[width=0.4\linewidth]{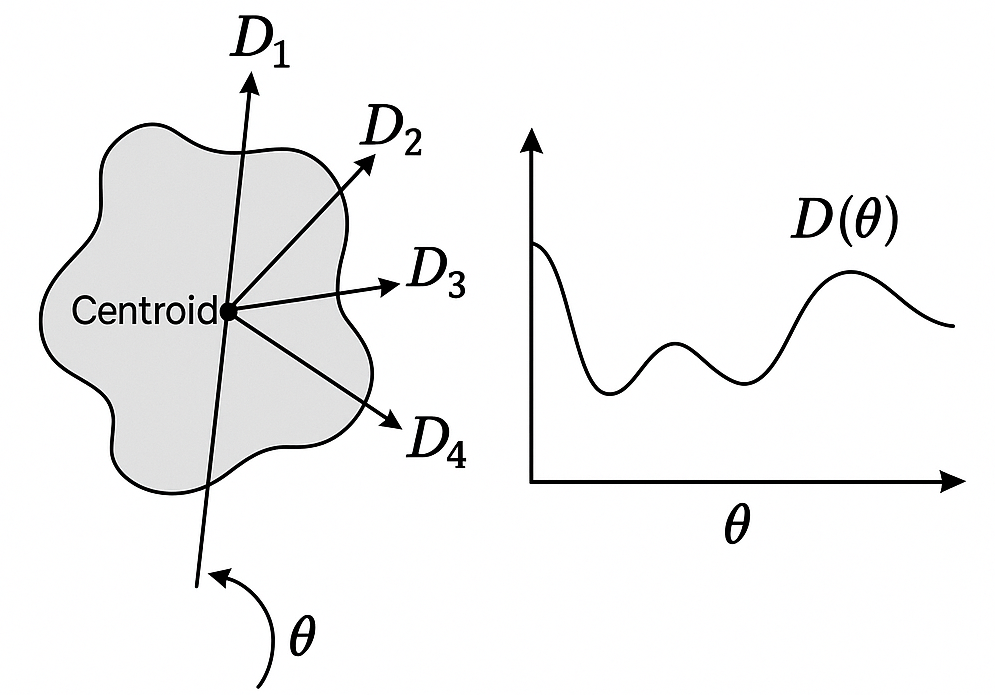}
     \caption{Left: An irregular 2D shape with its centroid marked. Radial lines drawn at various angles \(\theta\) extend from the centroid to the boundary, each labeled with a distance \(D_i\). Right: A plot of distance \(D(\theta)\) versus angle  \(\theta\), showing how the CDF captures shape variations around the perimeter.}
     \label{fig:cdf}
 \end{figure}

The \textit{centroid distance function} \citep{zhang2019chapter6} \(D(\theta)\) captures a shape's radial profile by measuring, for each direction \(\theta\), the distance from the object's centroid out to its boundary. After computing the centroid \((x_c, y_c)\), one casts a ray at angle \(\theta\) and finds the largest radius \(r\) such that the point  \((x_c+r\cos(\theta), y_c+r\sin(\theta))\) still lies inside the shape; see Figure \ref{fig:cdf}. 
In numerical implementations, this continuous function is sampled at equally spaced angles \(\theta_k=2\pi k/N\) for \(k=0, 1, \ldots, N-1\). At each \(\theta_k\), the boundary intersection  \((x(\theta_k), y(\theta_k))\) yields the discrete radius \(D_k=\sqrt{(x(\theta_k)-x_c)^2+(y(\theta_k)-y_c)^2}\).  Plotting \(D_k\)  against \(\theta_k\)  produces a signature that remains unchanged under translation -- since it is centered on the centroid -- and shifts cyclically under rotation. To eliminate size dependence, one divides \(D(\theta)\) by its mean value \(\bar{D}=\frac{1}{2\pi}\int_{2\pi}D(\theta)d\theta\), yielding the normalized profile \(\hat{D}(\theta)=D(\theta)/\bar{D}\). Finally, expressing \(D(\theta)\) as a Fourier series, \(D(\theta)=\sum_{n=-\infty}^{\infty} c_n e^{in\theta}\), and using the magnitudes \(|c_n|\) produces a compact, rotation- and scale-invariant set of descriptors, ideal for comparing arbitrary planar shapes.

\begin{figure}[!ht]
     \centering
     \includegraphics[width=0.6\linewidth]{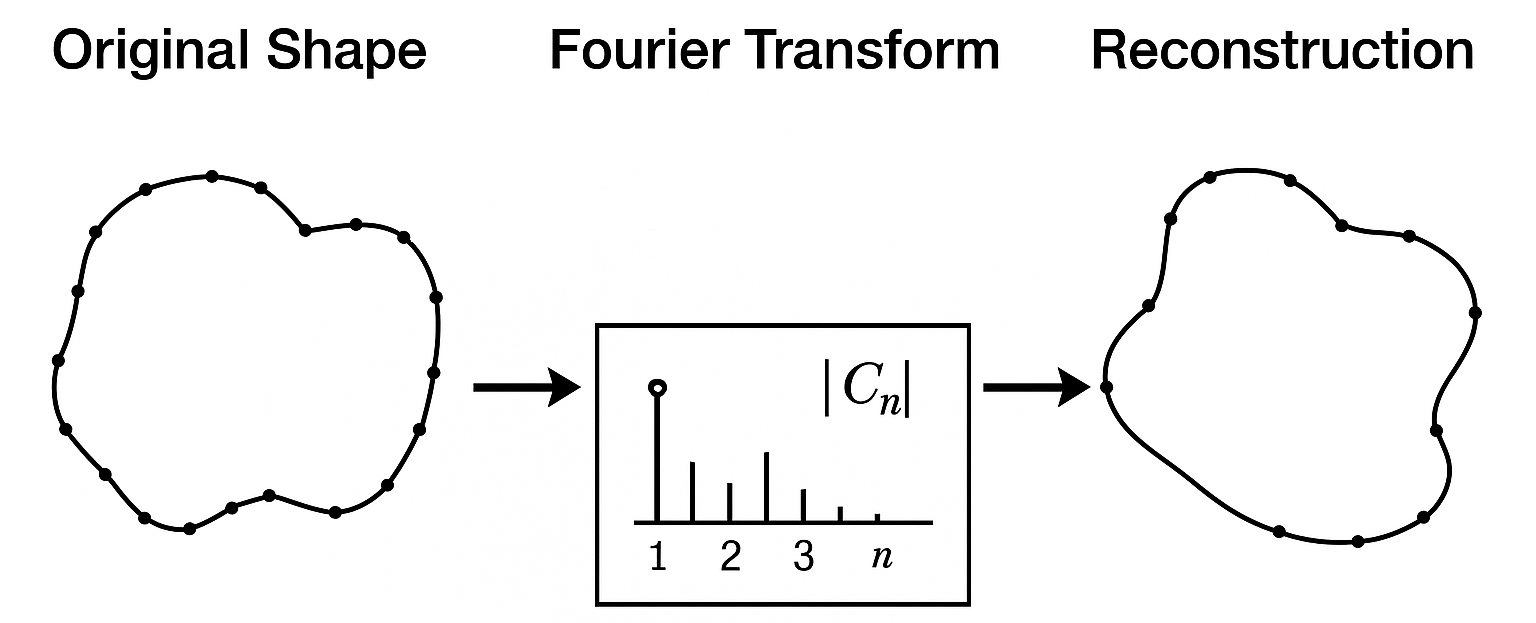}
     \caption{Original Shape (left): The closed contour sampled as complex points \(z(t)=x(t)+iy(t)\). Fourier Transform (center): The formula \(C_n=\frac{1}{N}\sum z(t)e^{-i2\pi nt/N}\) above a stem plot of coefficient magnitudes \(|C_n|\). Reconstruction (right): The approximated contour \(z'(t)=\sum C_ne^{i2\pi nt/N}\) showing the effect of using a limited number of harmonics.}
     \label{fig:fourier-descriptor}
 \end{figure}

\textit{Fourier descriptors} utilize the discrete Fourier transform to capture a shape's frequency components, enabling rotation invariance \citep{burger2013chapter6}. Figure \ref{fig:fourier-descriptor} illustrates the steps of generating Fourier descriptors. 
A closed planar contour can be treated as a periodic complex‐valued function \[z(t) = x(t) + i\,y(t), \quad t = 0,1,\dots,N-1,\] where \((x(t),y(t)) \) are the sampled boundary points in order. Taking its discrete Fourier transform gives the complex Fourier descriptors \[C_n = \frac{1}{N}\sum_{t=0}^{N-1} z(t)\,e^{-i2\pi n t / N},  \quad n = -\tfrac{N}{2},\dots,\tfrac{N}{2}-1.\] Because \(C_0\) is simply the centroid of the shape (translation term), one attains translation invariance by discarding or by zero‐centering: \(\tilde C_n = C_n - \delta_{n,0}\,C_0\).  To achieve scale invariance, divide every descriptor by the magnitude of the first harmonic (often \(|\tilde C_1|\): \(\hat C_n = \frac{\tilde C_n}{\lvert \tilde C_1\rvert}\).  Under a rigid rotation of the shape by angle \(\theta\), each descriptor picks up the same phase factor \(e^{i n\theta}\), so taking magnitudes \(F_n = \bigl\lvert \hat C_n \bigr\rvert\) yields rotation invariance. Finally, a reduced set of \(2K\) descriptors \(\{F_{-K}, \ldots, F_{-1}, F_1, \ldots, F_K\}\) provides a compact feature vector. The original contour can be approximately reconstructed by the inverse transform truncated to these harmonics: \(z'(t) = \sum_{n=-K}^{K} C_n\,e^{\,i2\pi n t / N}\). These Fourier descriptors thus form translation-, scale-, and rotation-invariant signatures well suited for comparing and classifying arbitrary shapes.

\textit{Zernike moments} \citep{zhang2019chapter6} project a gray‐level image \(f(x,y)\) onto an orthogonal basis of complex polynomials defined over the unit disk. First, one introduces the Zernike polynomial of order 
\(n\) and repetition \(m\) (where \(n\geq 0\), \(|m|\leq n\) and \(n-|m|\) is even) in polar coordinates \((r, \theta)\) by \[V_n^m(r,\theta)=R_n^{|m|}(r)\,e^{i m\theta},\] where the real radial polynomial is \[
R_n^{|m|}(r)
=\sum_{s=0}^{\frac{n-|m|}{2}}
\frac{(-1)^s(n-s)!}{s!\,\bigl(\frac{n+|m|}{2}-s\bigr)!\,\bigl(\frac{n-|m|}{2}-s\bigr)!}
\,r^{\,n-2s},
\]
ensuring orthogonality on the unit disk via
\[
\iint_{r\le1}
V_n^m(r,\theta)\,\overline{V_{n'}^{m'}(r,\theta)}\;r\,dr\,d\theta
=\frac{\pi}{n+1}\,\delta_{n,n'}\,\delta_{m,m'}.
\]
The Zernike moment \(A_n^m\) of the image is then defined by the inner product
\[
A_n^m
=\frac{n+1}{\pi}
\iint\limits_{x^2+y^2\le1}
f(x,y)\,\overline{V_n^m\bigl(r(x,y),\theta(x,y)\bigr)}\;dx\,dy,
\]
where \(r=\sqrt{x^2+y^2}\) and \(\theta=\mathrm{atan2}(y, x)\). Because the basis functions are orthogonal, each coefficient  \(A_n^m\) captures independent shape information. Translation invariance is achieved by centering the image’s region of interest within the unit disk before computation. Scale invariance follows from normalizing \(r\) by the maximum radius, and rotation of the image by angle \(\theta\) simply multiplies every moment by \(e^{-im\theta}\) so the magnitudes \(|A_n^m|\) form rotation‐invariant descriptors.

\subsection{Clustering Methods}
To uncover the underlying variability in powder-particle shapes, we introduce and compare three unsupervised clustering approaches, each built upon a distinct representation of morphology. The first pipeline leverages an O(2)-invariant variational autoencoder: binary particle masks are encoded through a sequence of steerable convolutions that respect rotations and reflections, yielding a 64-dimensional latent code invariant to orientation. A conventional convolutional decoder then reconstructs each mask, and we apply clustering methods (k-means and Gaussian mixture model (GMM)) on the latent representations. In the second approach, we distill each particle's geometry into a suite of invariant descriptors -- centroid distance functions, Fourier descriptors, and Zernike moments -- and, when necessary, reduce their dimensionality via principal component analysis (PCA). This descriptor-based pipeline similarly applies both hard (k-means) and soft (GMM) clustering algorithms, allowing us to contrast the convenience and speed of Euclidean-distance grouping (i.e., k-means) with the flexibility of elliptical cluster boundaries that can accommodate overlapping or elongated shape classes (i.e., GMM). The third strategy treats each centroid distance function as a continuous curve in angular space and employs GPmix, a functional data clustering method, to cluster the radial profile functions directly. Below, we provide a concise overview of each pipeline.

\begin{figure}[!ht]
    \centering
    \includegraphics[width=0.7\linewidth]{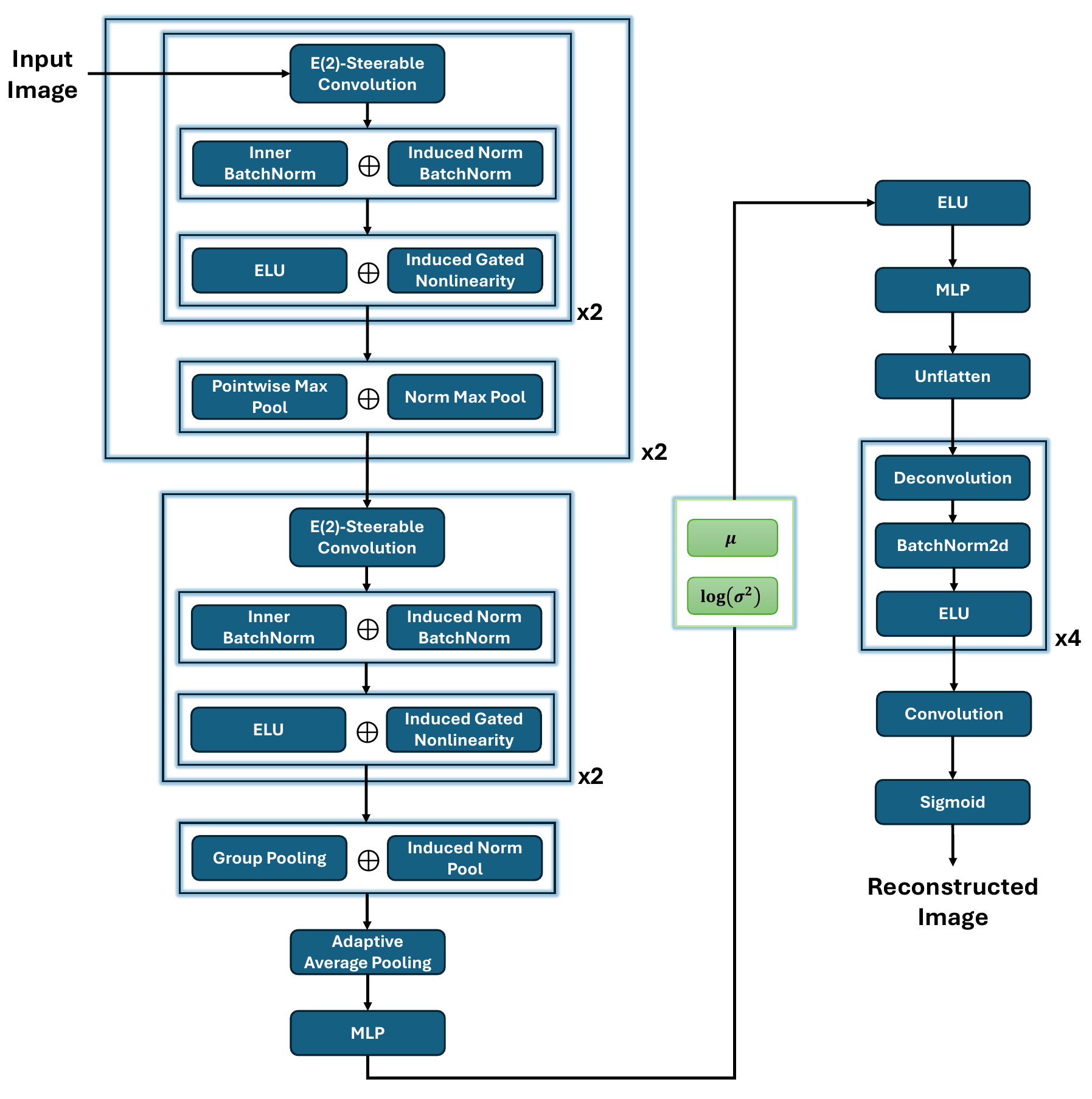}
    \caption{Schematic of the O(2)-invariant VAE. $\mathbf{\oplus}$ denotes the concatenation of output feature fields. The encoder uses six E(2)-steerable convolutions (with batch normalization and nonlinearity), inserts equivariant pooling after layers 2 and 4, and applies global group pooling after layer 6. The encoder's MLP block is a four-stage sequence: a linear projection, a 1D batch-normalization layer, an ELU activation, and a final linear layer. The resulting embeddings are passed to a decoder built from a standard convolutional architecture.}
    \label{fig:o2vae-arch}
\end{figure}

Autoencoders offer an elegant solution for learning compact, unsupervised representations by mapping high-dimensional inputs into a low-dimensional latent space and then reconstructing the originals. In the context of shape analysis, however, it is essential that these latent representations remain unchanged under shape-preserving transformations like planar rotations and reflections -- symmetries that ordinary convolutional networks do not guarantee. E(2)-equivariant neural networks provide this capability by construction, ensuring that learned features transform consistently under rotations and reflections \citep{e2cnn, cohen2019general}. To our knowledge, such equivariant (or invariant) architectures have not previously been applied to particle- or powder-shape analysis in materials science; existing relevant works primarily target atomic or microstructural symmetry -- for example, SE(3)- or E(3)-equivariant graph neural networks for crystal property prediction \citep{kaba2022equivariant, pakornchote2023straintensornet} -- rather than morphological shape representations. To address this gap, we extend the orientation-invariant variational autoencoder framework in \citep{burgess2024orientation} with the theory of E(2)-equivariant steerable CNNs \citep{e2cnn} to build a VAE whose encoder is intrinsically invariant to the full orthogonal group O(2). Figure \ref{fig:o2vae-arch} illustrates the full architecture of the encoder–decoder model. 
\begin{itemize}
    \item Our encoder adopts the ``gated induced irrep O(2) model'', which demonstrated top performance among O(2)-invariant architectures in empirical tests \citep{e2cnn}. Each of its six convolutional blocks applies steerable filters that respect both rotations and reflections; batch normalization follows, and nonlinearities are applied separately to trivial and nontrivial irreducible representation channels, with ELU activations on the former. Kernel sizes shrink from $7\times7$ in the first layer (padding = 1) to $5\times5$ thereafter (padding = 2), while channel counts increase progressively (6 $\rightarrow$ 9 $\rightarrow$ 12 $\rightarrow$ 12 $\rightarrow$ 19 $\rightarrow$ 25). To downsample without breaking equivariance, we interleave equivariant pooling layers after the second and fourth convolution blocks. After the sixth convolution block, a global group-pooling operation outputs features which are invariant over all O(2) transformations -- that is, rotations and reflections. Finally, these invariant features are passed through a multilayer perception (MLP) block yielding latent embeddings that are agnostic to orientation.
    \item The decoder retains a standard convolutional structure, reconstructing binary masks from the invariant representations. Two fully connected layers with ELU activations expand the latent vector (the MLP block), which is then reshaped into a coarse spatial map (the unflatten layer). Four successive transposed convolutions (each with $4\times4$ kernels, stride = 2, padding = 1) upsample this map while reducing channel depth (32 $\rightarrow$ 16 $\rightarrow$ 8 $\rightarrow$ 1). A final $3\times3$ convolution smooths the output, and a sigmoid activation constrains pixel values to [0,1], mitigating reconstruction artifacts.
    \item By marrying group-representation theory with variational inference, this O(2)-invariant VAE ensures that latent codes capture only intrinsic morphological features of powder particles, discarding arbitrary orientations. In practice, this symmetry-aware design delivers both greater sample efficiency and provably consistent embeddings, laying a rigorous foundation for downstream tasks such as clustering or generative modeling of complex shapes.
\end{itemize}

In the shape-descriptor pipeline, each binary mask is represented by an invariant feature vector -- either a centroid distance function (sampled over a fixed grid), a set of Fourier descriptors, or Zernike moments (as detailed in Section \ref{shape-descriptors}). When these descriptors are high-dimensional, we apply PCA to project them onto a lower-dimensional, more tractable subspace. Clustering is then performed on each set of feature descriptors using both k-means and GMM.

Finally, because centroid distance functions define smooth closed contours on $[0, 2\pi)$, we also treat them as functional data and cluster them using GPmix \citep{akeweje2024learning}. GPmix assumes that the observed curves arise from a finite mixture of Gaussian processes and avoids fitting this infinite-dimensional model directly. Instead, each radial function is projected onto a small number of randomly generated basis functions; for each projection, the resulting scalar coefficients are modeled with a univariate Gaussian mixture, yielding a “base” clustering. An ensemble-clustering step then combines these base clusterings into a single consensus partition. Applied to our centroid-distance profiles, GPmix directly clusters the radial functions themselves rather than hand-crafted descriptors, providing a scalable and statistically principled way to detect distinct shape groups while limiting information loss relative to descriptor-based methods.

Because no ground-truth labels are available for our dataset, clustering quality is assessed using internal validity indices that directly quantify the structure of the partitions. For the k-means method, the number of clusters $K$ was chosen by evaluating Silhouette scores over a candidate range of $K=2$ to $9$ and selecting the value that maximized this criterion. For the GMM, the number of mixture components was determined using the Bayesian Information Criterion (BIC), which penalizes overly complex models and therefore discourages overfitting. GPmix similarly uses its built-in BIC-based mechanism to infer the number of clusters from the functional data. After these primary selections, the resulting partitions were further examined using the Davies–Bouldin (DB) index, which quantifies the balance between within-cluster compactness and between-cluster separation, and the Calinski–Harabasz (CH) score, which measures the ratio of between-cluster to within-cluster dispersion. When the Silhouette or BIC curves did not exhibit a single clear optimum -- for example, when several adjacent values of $K$ performed similarly -- DB and CH served as secondary diagnostics to confirm a stable and physically reasonable choice of $K$. These same indices also provide a common basis for comparing clustering quality across pipelines, where lower DB and higher CH values indicate better separation of morphological subgroups. We also adopted circularity and aspect-ratio, two widely used morphological descriptors in the materials science literature, as external validation proxies for assessing the clusterings.

\section{Results}\label{Results}
The clustering methods employed in this study are organized into three categories, and the results are presented accordingly in this section. The source code for implementing these methods is publicly available at \url{https://github.com/EAkeweje/Powder-Morphology-Analysis.git}.

\subsection{Clustering of VAE Embeddings}
\begin{figure}[!h]
    \begin{subfigure}[c]{\linewidth}
    \centering
    \includegraphics[width=0.6\linewidth]{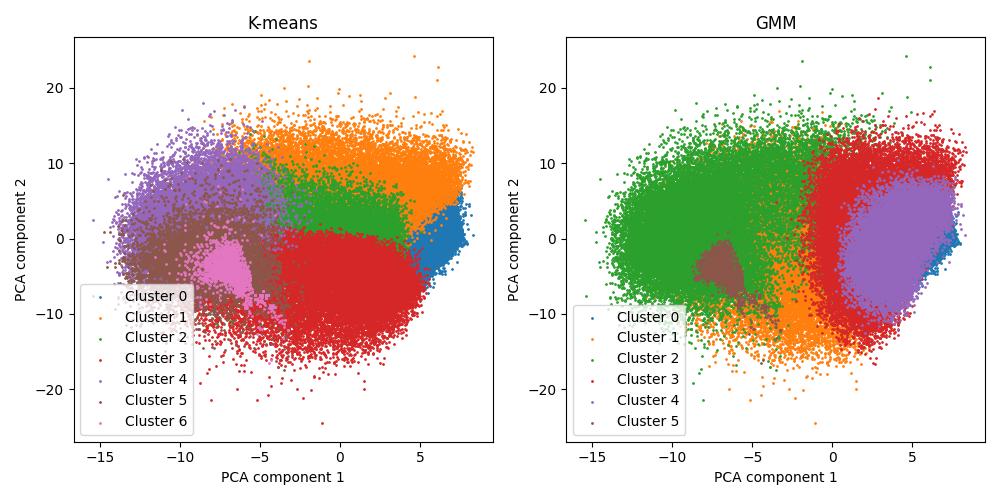}
    \caption{Two-dimensional PCA projection of the 64-dimensional VAE latent codes, with each point colored by its k-means (left) or GMM (right) assignment.}
    \label{fig:vae-labels}
    \end{subfigure}
    
    \begin{subfigure}[c]{0.48\linewidth}
    \includegraphics[width=\linewidth]{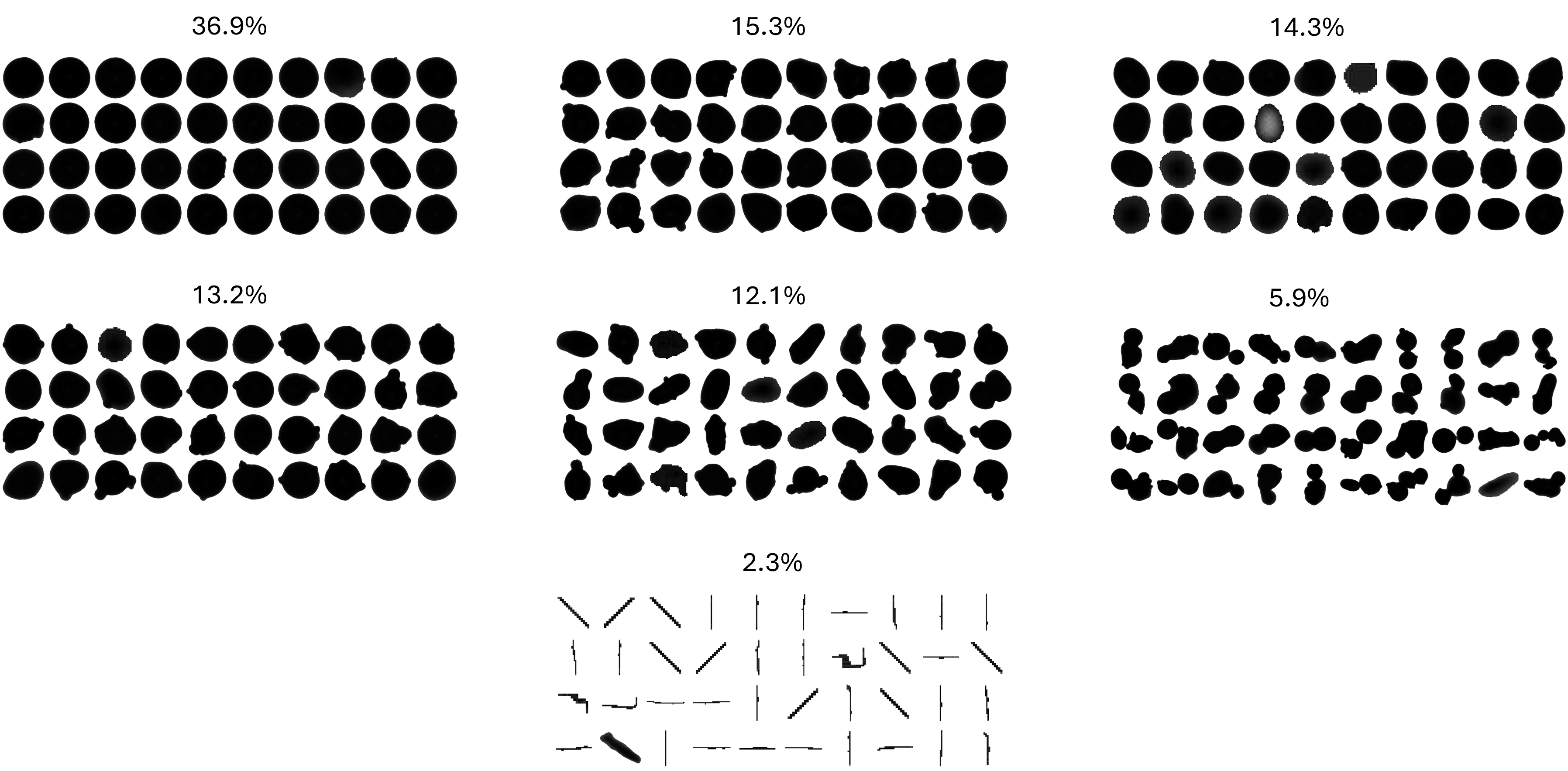}
    \caption{Representative binary masks from each of the seven k-means clusters, ordered by decreasing cluster size. Percentages showing each cluster’s share of the dataset. Clusters 2 ($14.3\%$) to 7 ($2.3\%)$ manifest progressively more complex shapes: slight contour irregularities, satellite-bearing forms, agglomerated clumps, fused/deformed morphologies, and highly fragmented debris. }
    \label{fig:vae-kmeans}
    \end{subfigure}
    \hfill
    \begin{subfigure}[c]{0.48\linewidth}
    \centering
    \includegraphics[width=\linewidth]{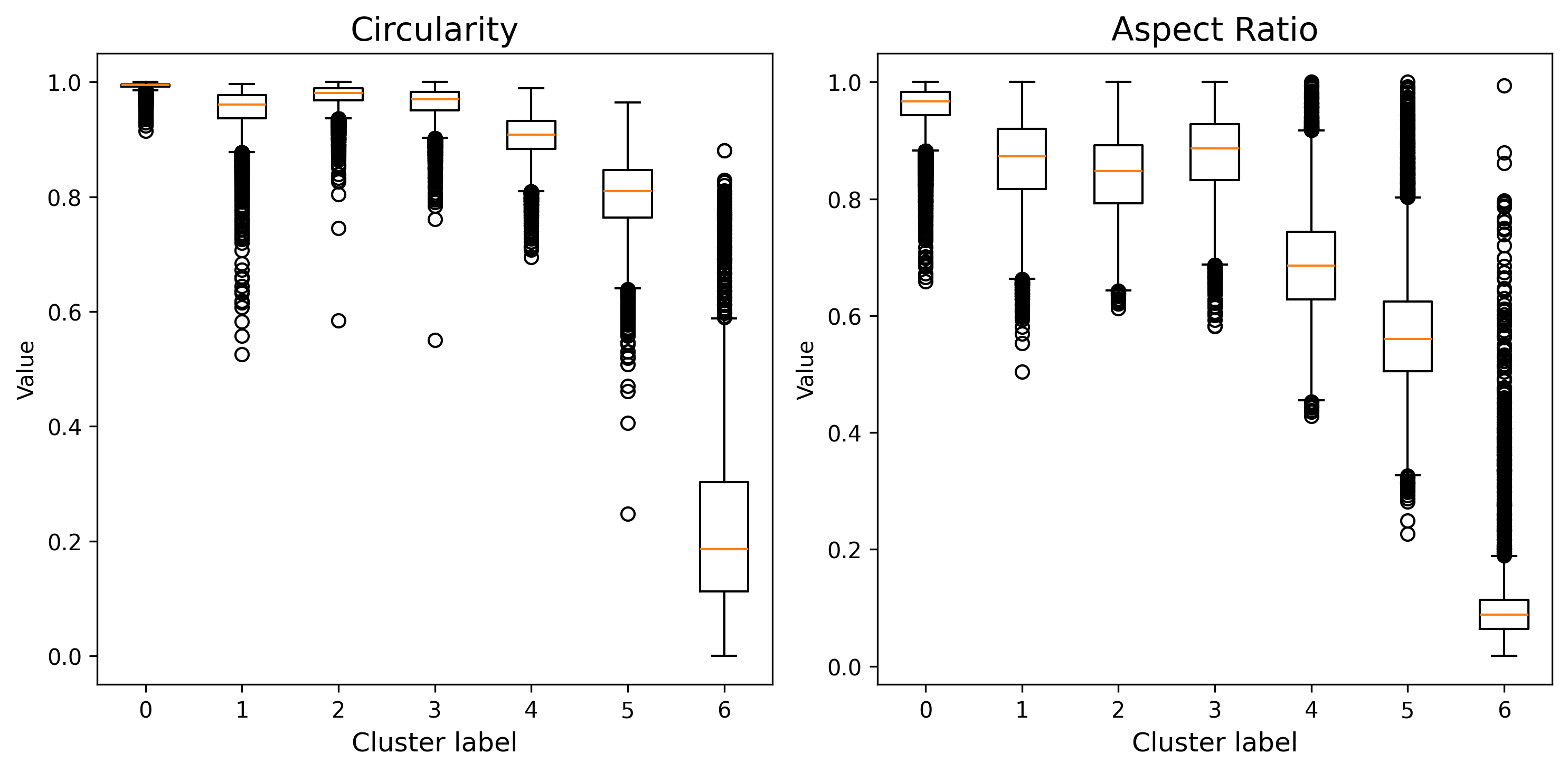}
    \caption{The circularity and aspect-ratio distributions across the seven k-means clusters. Particles in the second to fourth clusters have similar morphology.}
    \label{fig:vae_kmeans_boxplot}
    \end{subfigure}
    
    \begin{subfigure}[c]{0.48\linewidth}
    \includegraphics[width=\linewidth]{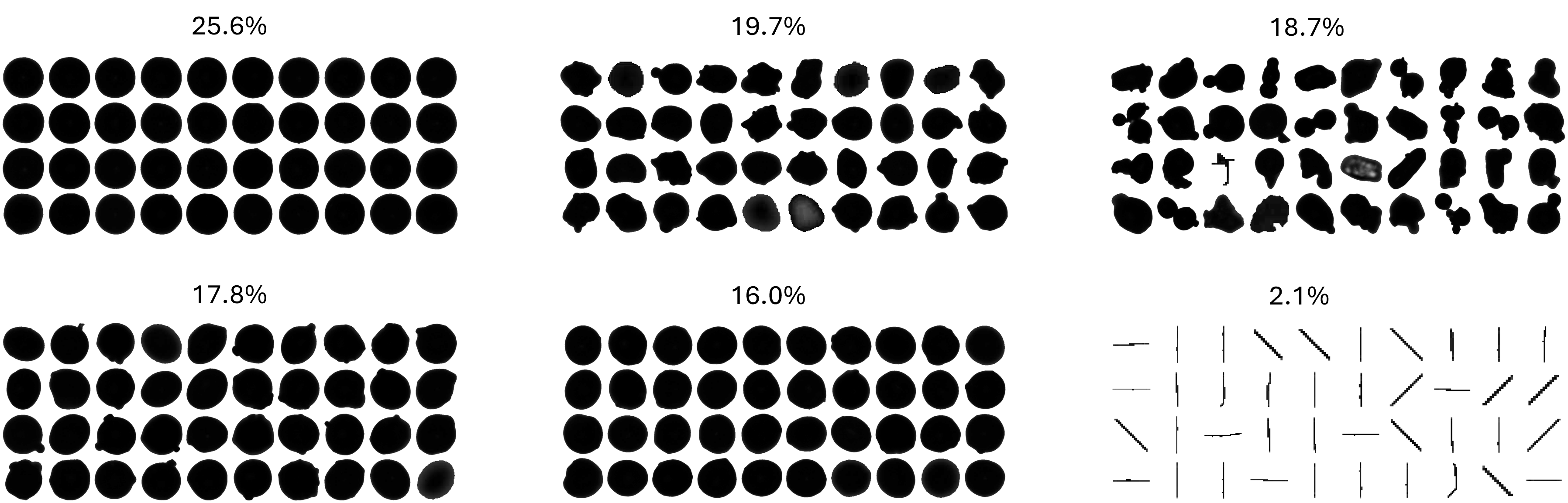}
    \caption{The first GMM cluster (25.6\% of particles) captures ideal spheres most cleanly, the third cluster (18.7\%) groups highly distorted/agglomerated particles, and cluster 6 captures the most fragmented debris.}
    \label{fig:vae-gmms}
    \end{subfigure}
    \hfill
    \begin{subfigure}[c]{0.48\linewidth}
    \centering
    \includegraphics[width=\linewidth]{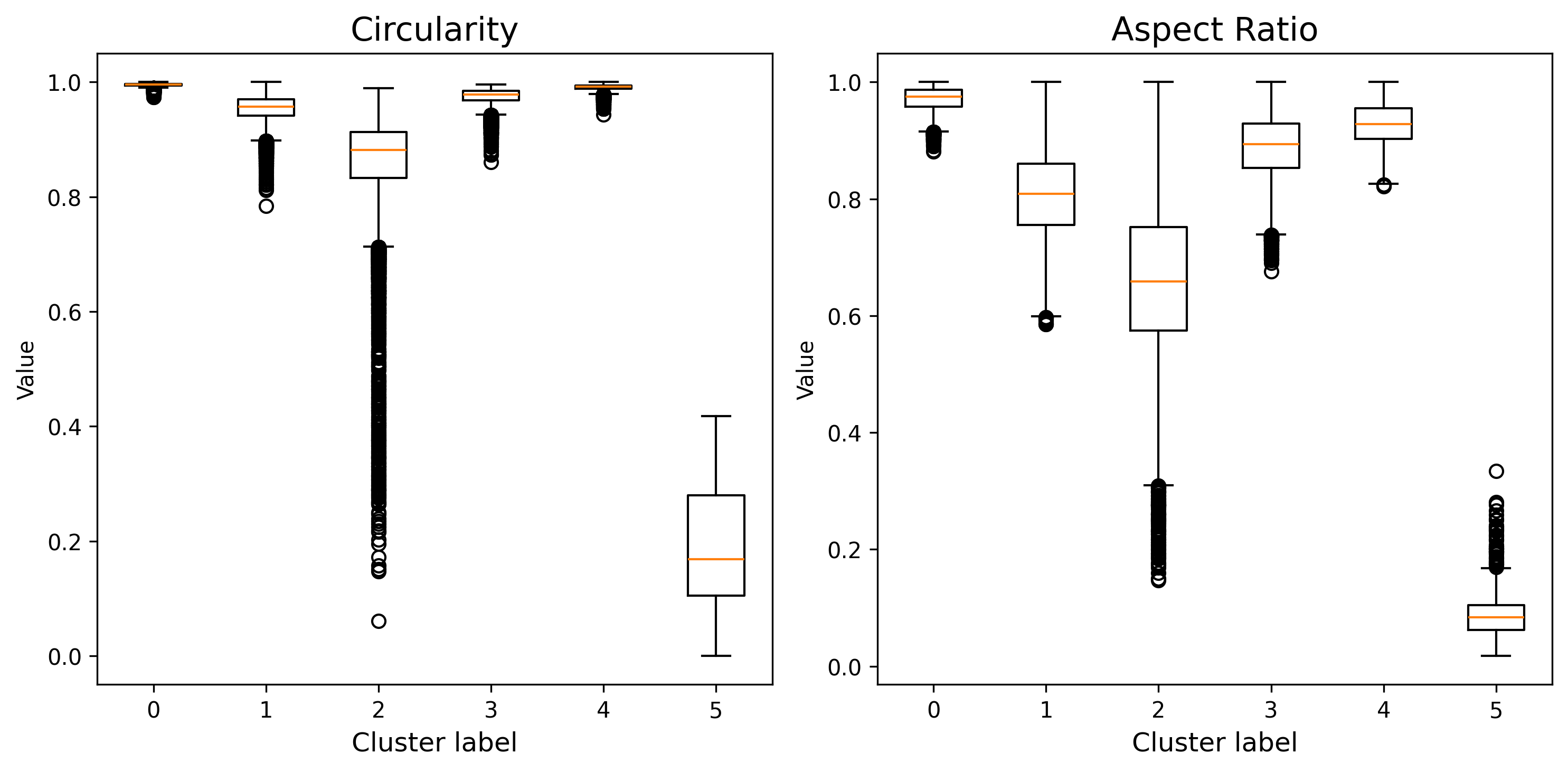}
    \caption{The circularity and aspect-ratio distributions across the six GMM clusters, where the morphology of the sixth cluster is distinct from the others.}
    \label{fig:vae_gmm_boxplot}
    \end{subfigure}
    
    \caption{Clustering via VAE embedding. }
    \label{fig:vae-samples}
\end{figure}

To evaluate the representational capacity of the VAE, we trained models with latent dimensions of 16, 32, 64, and 128 and compared their reconstruction performance. A 64-dimensional latent space achieved the lowest reconstruction loss without the training instability or signs of over-parameterization observed at higher dimensions, and we therefore adopt this latent size in the final model. Because the network is invariant to in-plane rotations and reflections, reconstructed masks may be misaligned with the original inputs; before computing pixel-wise reconstruction error, we therefore estimate the optimal rigid alignment between input and reconstruction using phase correlation and evaluate the error in this aligned frame.

We optimized using Adam (learning rate \(10^{-3}\), \(\beta_1=0.5\), \(\beta_2=0.9\)), decaying the learning rate by 10\% every 10 epochs. Training used mini-batches of 64, with data split 70:15:15 into training, validation, and test sets. We ran for up to 200 epochs, stopping early if the validation loss failed to improve for 20 consecutive epochs. After convergence, we passed every mask through the encoder and retained the posterior means \(\boldsymbol\mu_{\boldsymbol\phi}(x)\) as our 64-dimensional embeddings.

We then applied k-means and GMM to these embeddings and visualized the results via two-dimensional PCA projections (Figure \ref{fig:vae-labels}). Under k-means, seven clusters were identified; the clusters occupy moderately distinct regions of the latent space, yielding a DB index of 1.7638 and a CH score of 150,88 -- metrics that together indicate compact, well-separated groups. To reduce the computational cost of inverting 64×64 covariance matrices for the GMM, we first projected the 64-dimensional latent space onto 20 principal components. This projection retains 98.9\% of the variance in the original latent embeddings, ensuring that the dominant morphological structure learned by the VAE is preserved. The GMM identifies six clusters, with a higher DB index of 2.9105 and a lower CH score of 9,827, reflecting more ambiguous cluster boundaries.

The exemplar masks in Figures \ref{fig:vae-kmeans} and \ref{fig:vae-gmms} underscore the distinct ways in which k-means and GMM capture particle morphology. In the k-means clustering, more than a third of the particles (36.9\%) form a tight, near-spherical group, while the remaining clusters progressively reflect increasing shape complexity: clusters 2 and 3 retain roughly circular shapes but show small edge indentations or missing ``chunks''; cluster 4 exhibits pronounced surface irregularities  --  bulges, concavities and uneven rims; cluster 5 consists of lens- or bean-shaped particles with a well-defined major axis; cluster 6 is dominated by fused ``peanut'' or multi-lobed agglomerates; and cluster 7 encompasses thin, elongated, needle-like fragments with high aspect ratios. This clear stratification of simple versus complex shapes attests to the VAE's ability to embed salient morphological features in a linearly separable latent space. In the GMM partition, the first cluster cleanly isolates ideal spheres (25.6 \%), while its third cluster (18.7 \%) largely recapitulates the deformed and agglomerated morphologies captured by k-means clusters 5 and 6. Compared to the k-means clusters, GMM clusters blend intermediate morphologies, yielding broader, more ambiguous boundaries among round, deformed, and fragmented particles. Both clustering methods consistently isolate a distinct cluster dominated by highly fragmented debris.

\subsection{Clustering of Extracted Morphometric Data}
\begin{figure}[!h]
    \begin{subfigure}[c]{\linewidth}
    \centering
    \includegraphics[width=0.6\linewidth]{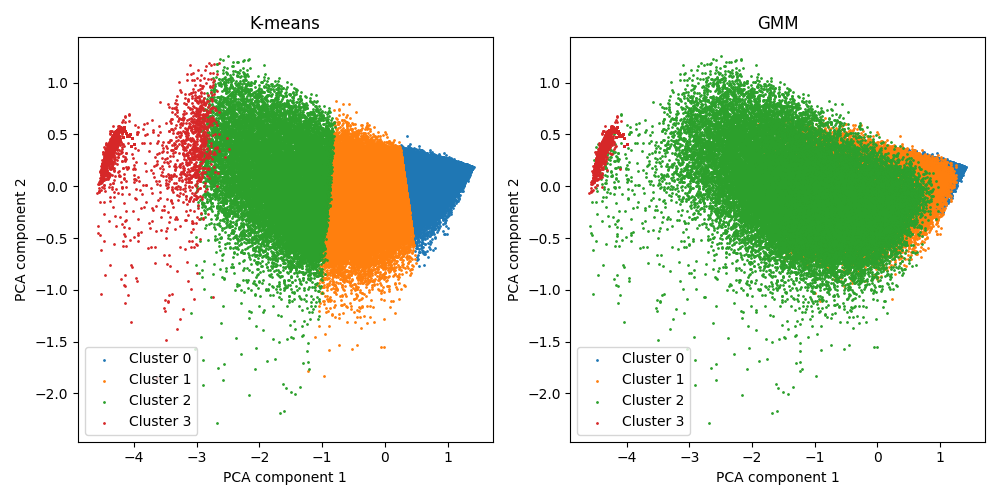}
    \caption{Two-dimensional PCA projection of the 100-dimensional CDF features colored by four k-means clusters (left) and GMM clusters (right). }
    \label{fig:cdf-labels}
    \end{subfigure}

    \begin{subfigure}[c]{0.48\linewidth}
    \includegraphics[width=\linewidth]{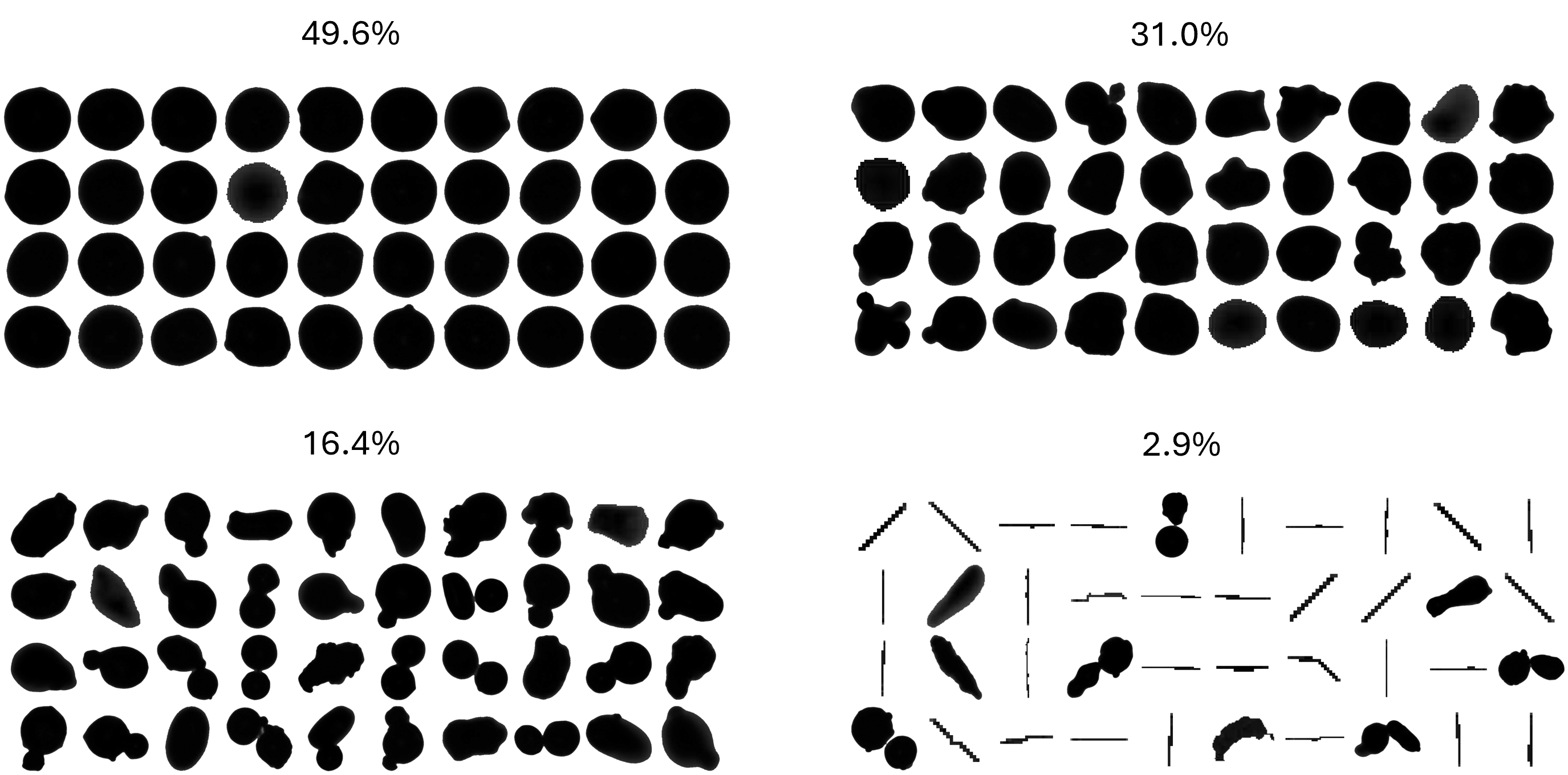}
    \caption{Representative particle shapes for each k-means cluster (based on CDF features), with percentages showing each cluster’s share of the dataset.}
    \label{fig:cdf-kmeans}
    \end{subfigure}
    \hfill
    \begin{subfigure}[c]{0.48\linewidth}
    \centering
    \includegraphics[width=\linewidth]{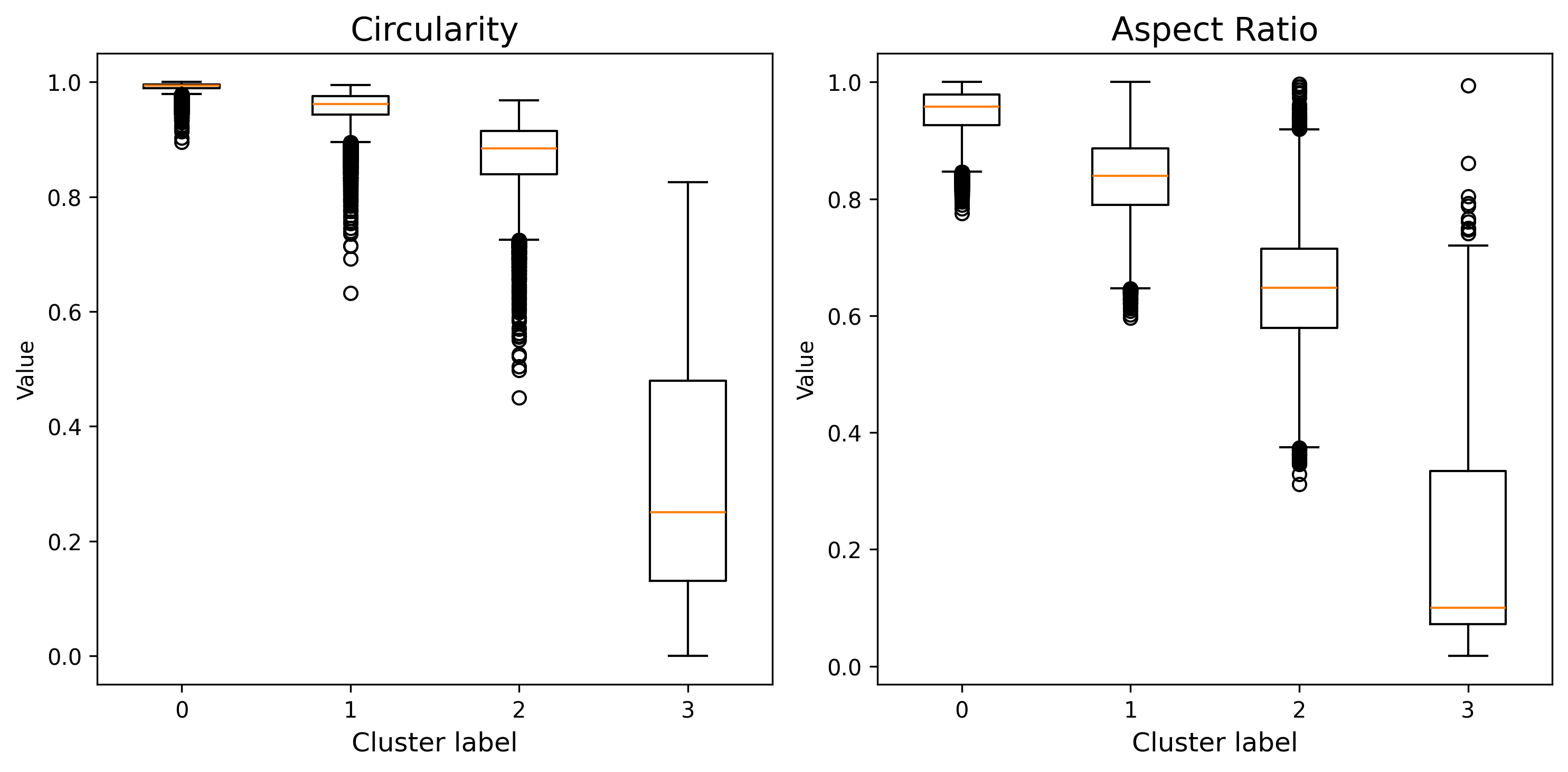}
    \caption{Boxplots of circularity (higher = more round) and aspect ratio (width/length) across the k-means clusters.}
    \label{fig:cdf_kmeans_boxplot}
    \end{subfigure}
    
    \begin{subfigure}[c]{0.48\linewidth}
    \includegraphics[width=\linewidth]{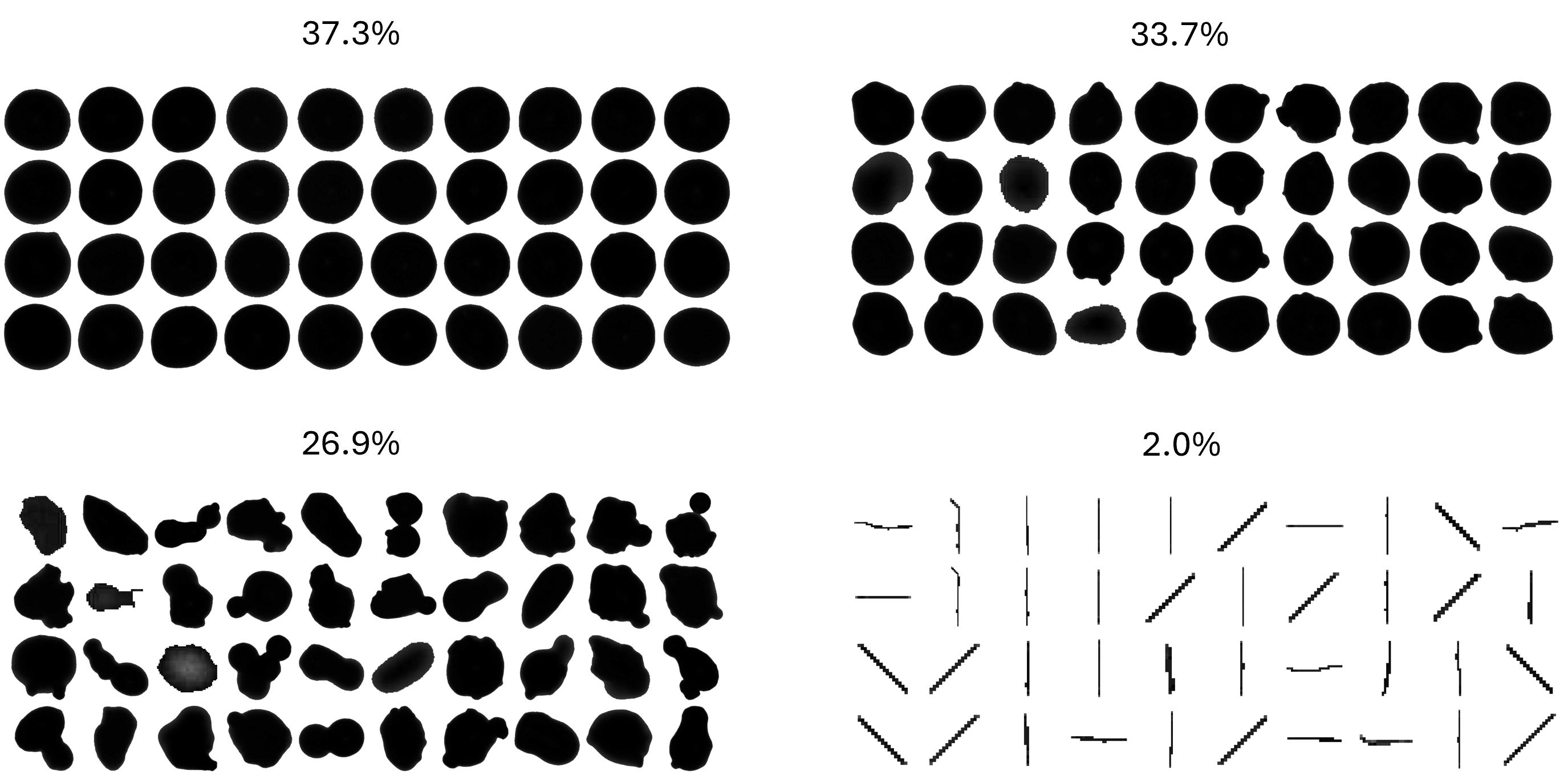}
    \caption{Representative particle shapes for each GMM cluster (based on CDF features), with percentages showing each cluster’s share of the dataset.}
    \label{fig:cdf-gmms}
    \end{subfigure}
    \hfill
    \begin{subfigure}[c]{0.48\linewidth}
    \centering
    \includegraphics[width=\linewidth]{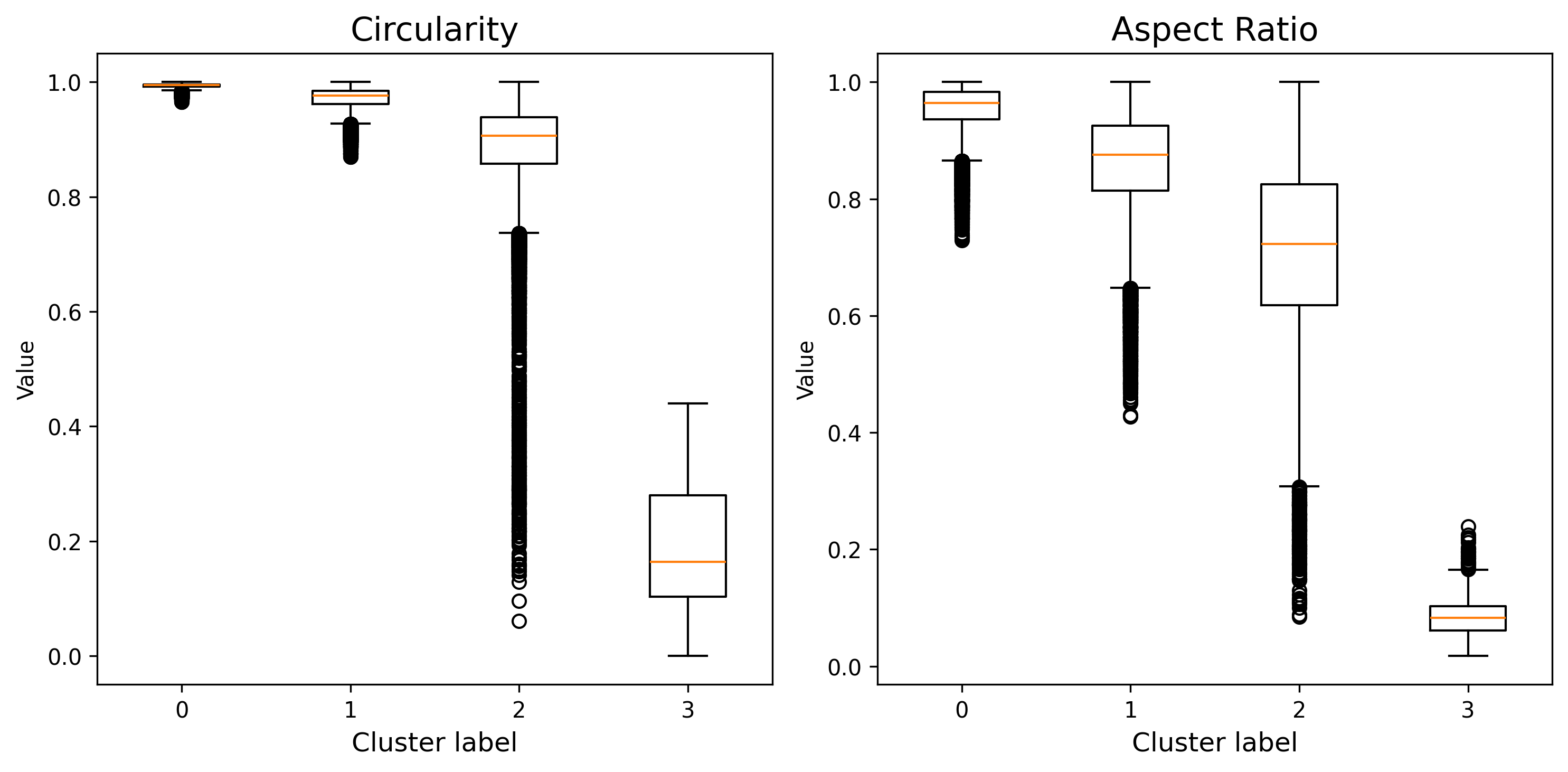}
    \caption{Boxplots of circularity and aspect ratio across the GMM clusters.}
    \label{fig:cdf_gmm_boxplot}
    \end{subfigure}
    
    \caption{Both clustering methods identify four clusters in the CDF-feature data.}
    \label{fig:cdf-all-results}
\end{figure}

\paragraph{Morphology analysis via CDF features} We first extracted 100-dimensional CDF descriptors. For k-means, clustering was performed directly in this feature space, whereas for GMM the descriptors were first reduced to 20 principal components to alleviate the computational cost of inverting large covariance matrices. This 20-dimensional projection preserves 99.7\% of the variance in the original CDF descriptors, ensuring that essentially all relevant morphological variability is retained. The clustering results are summarized in Figure \ref{fig:cdf-all-results}. In the two-dimensional PCA scatterplot (Figure \ref{fig:cdf-labels}), k-means slices the PCA cloud into four roughly equal ``vertical'' bands along PC1. By contrast, the GMM result uses overlapping Gaussian ellipsoids. Note that the orange (Cluster 1) points lie almost entirely beneath the green (Cluster 2) -- it effectively ``hides'' behind Cluster 2 in this projection. Likewise, the blue (Cluster 0) points are largely overlapped by the orange cluster and thus not directly visible in this projection.

Inspection of representative masks (Figures \ref{fig:cdf-kmeans} and \ref{fig:cdf-gmms}) confirms these patterns. Under k-means, the first panel (49.6\%) is populated almost entirely by near-perfect, disk-shaped particles -- smooth, circular silhouettes with minimal irregularity. In the second panel (31\%), the shapes remain broadly round but exhibit gentle undulations and slight indentations around their peripheries. In the third panel (16.4\%), those undulations become more pronounced and complex, with many particles showing multi-lobed or blob-like protrusions. Finally, the bottom-right panel is dominated by highly elongated, rod- or needle-like fragments, many appearing as thin sticks or connected bead-chains rather than compact blobs.  GMM’s Cluster 1 contains only the most nearly perfect disks, while its Cluster 2 picks up just the gently undulated forms. The third GMM cluster then isolates the more strongly lobed, blob-like particles, and the final cluster still captures the thin, rod- or needle-shapes. These nuances are echoed in the boxplots: the third GMM cluster's whiskers extend farther than those of k-means, while the final GMM cluster displays much shorter whiskers, indicating that its needle-like forms are more consistently elongated.

These qualitative insights are borne out by internal validation metrics. k-means achieves a DB index of 0.8513 and a CH score of 173,352, reflecting strong intra-cluster cohesion and clear inter-cluster separation. GMM, while better at capturing intra-cluster heterogeneity, exhibits a higher DB index of 1.0767 and a lower CH score of 80,323, indicating that its increased flexibility comes at the cost of cluster compactness and distinctiveness.  These metrics indicate that although GMM captures more nuanced shape variations, it sacrifices cluster compactness and separation in doing so. This trade-off may be attributable to the model’s flexibility, which enables it to model data with higher intra-cluster variance.

\begin{figure}[!h]
    \begin{subfigure}[c]{\linewidth}
    \centering
    \includegraphics[width=0.6\linewidth]{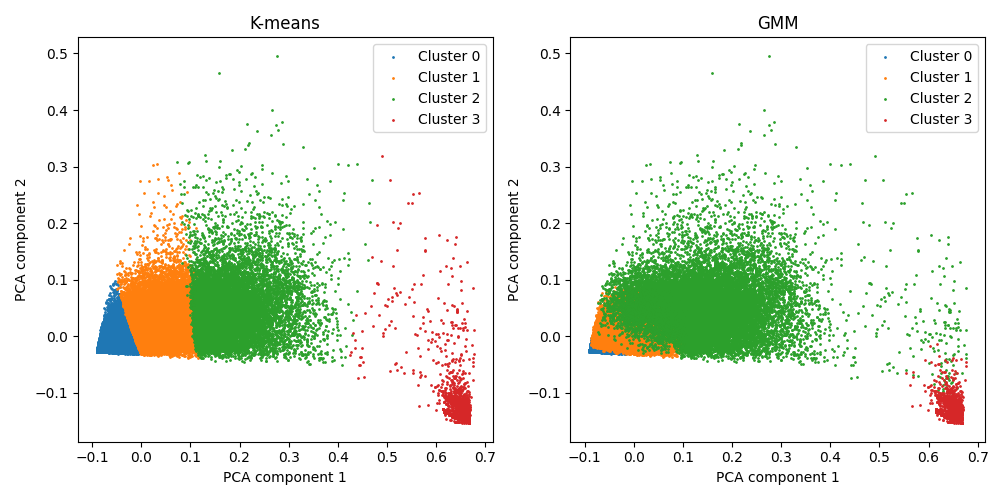}
    \caption{Two-dimensional PCA projection of the 10-dimensional FD features colored by four k-means clusters (left) and GMM clusters (right).}
    \label{fig:fd-labels}
    \end{subfigure}

    \begin{subfigure}[c]{0.48\linewidth}
    \includegraphics[width=\linewidth]{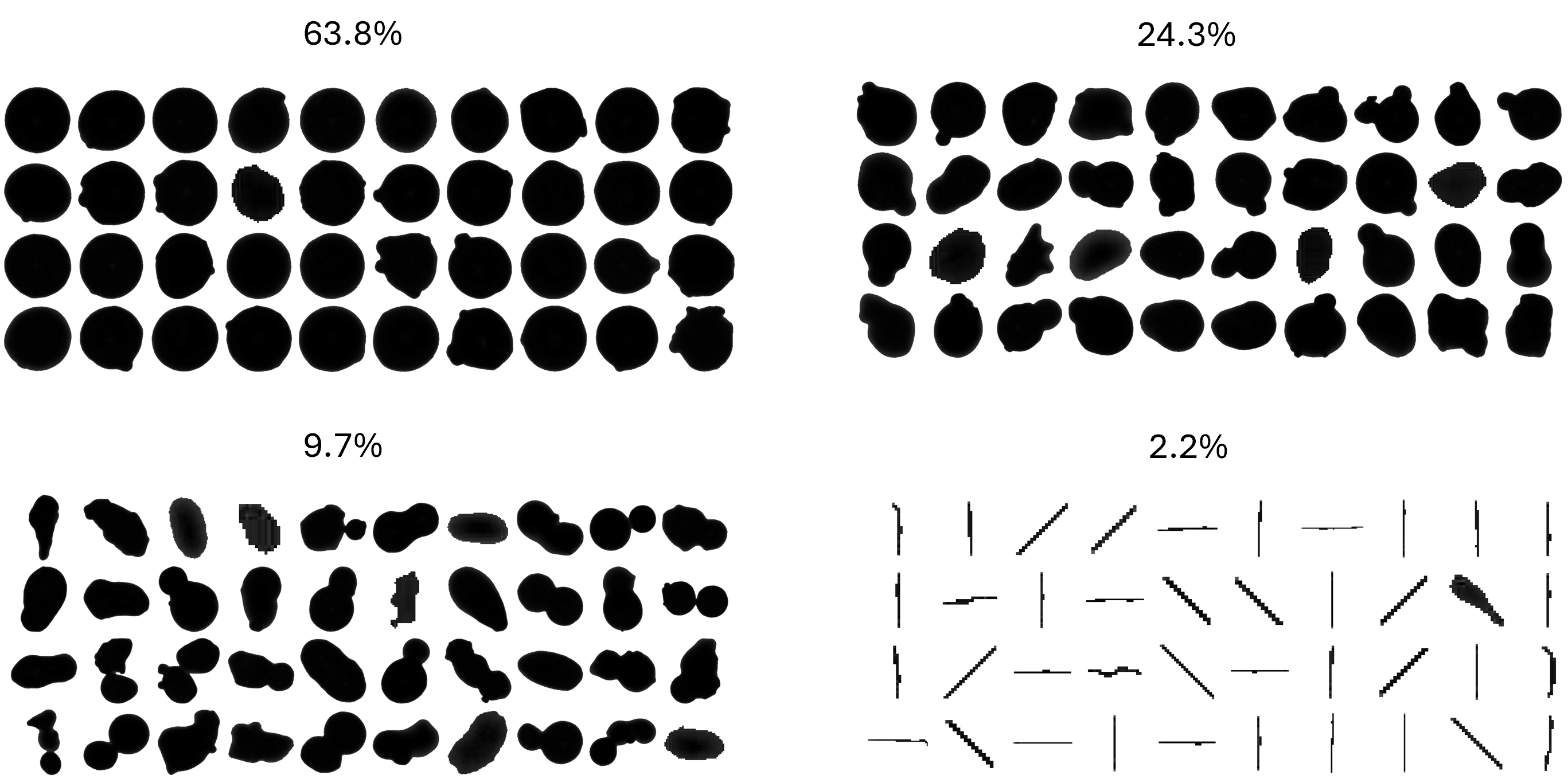}
    \caption{Representative particle shapes for each k-means cluster (based on FD features), with percentages showing each cluster’s share of the dataset.}
    \label{fig:fd-kmeans}
    \end{subfigure}
    \hfill
    \begin{subfigure}[c]{0.48\linewidth}
    \centering
    \includegraphics[width=\linewidth]{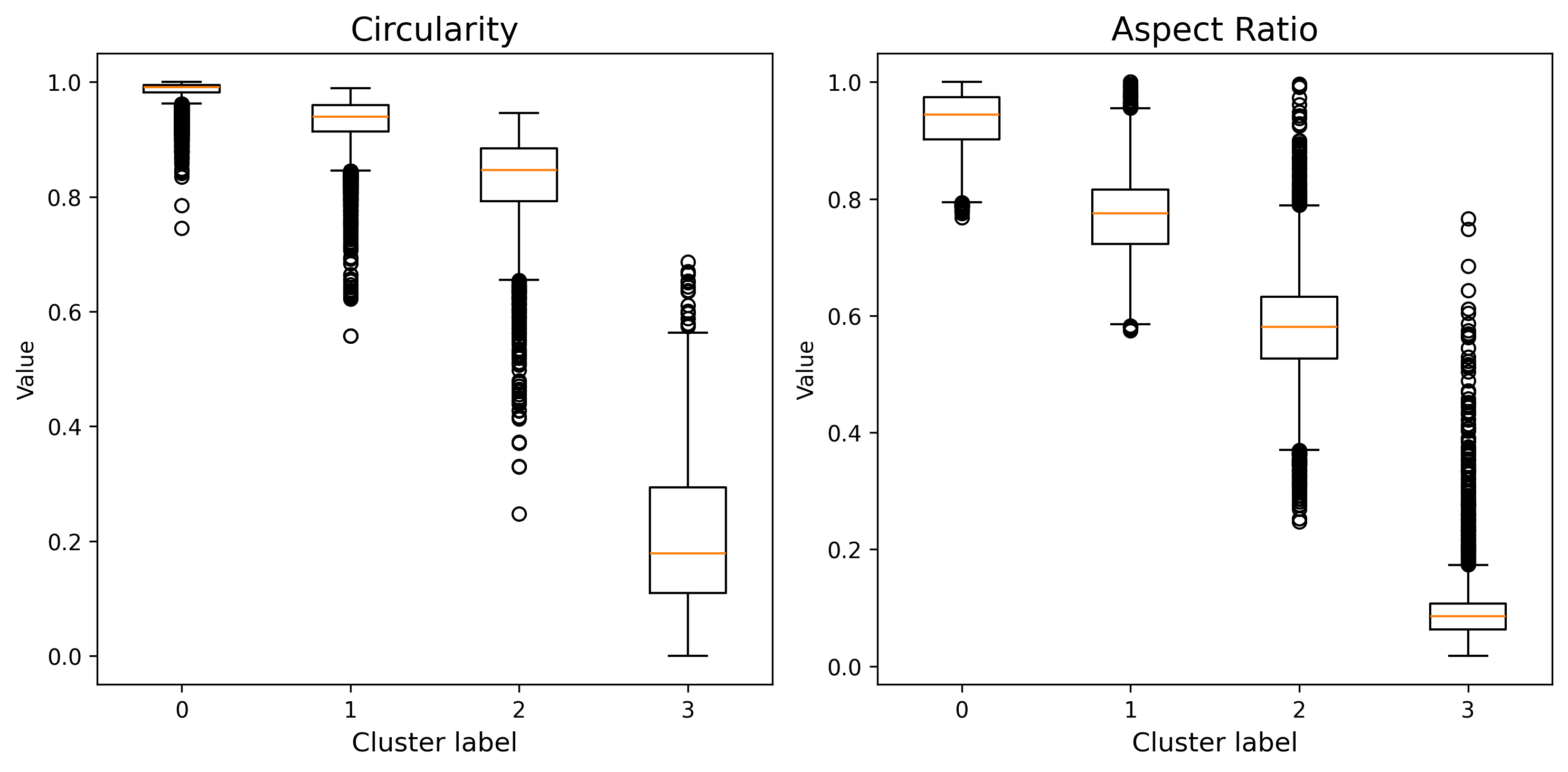}
    \caption{Boxplots of circularity (higher = more round) and aspect ratio (width/length) across the k-means clusters.}
    \label{fig:fd_kmeans_boxplot}
    \end{subfigure}
    
    \begin{subfigure}[c]{0.48\linewidth}
    \includegraphics[width=\linewidth]{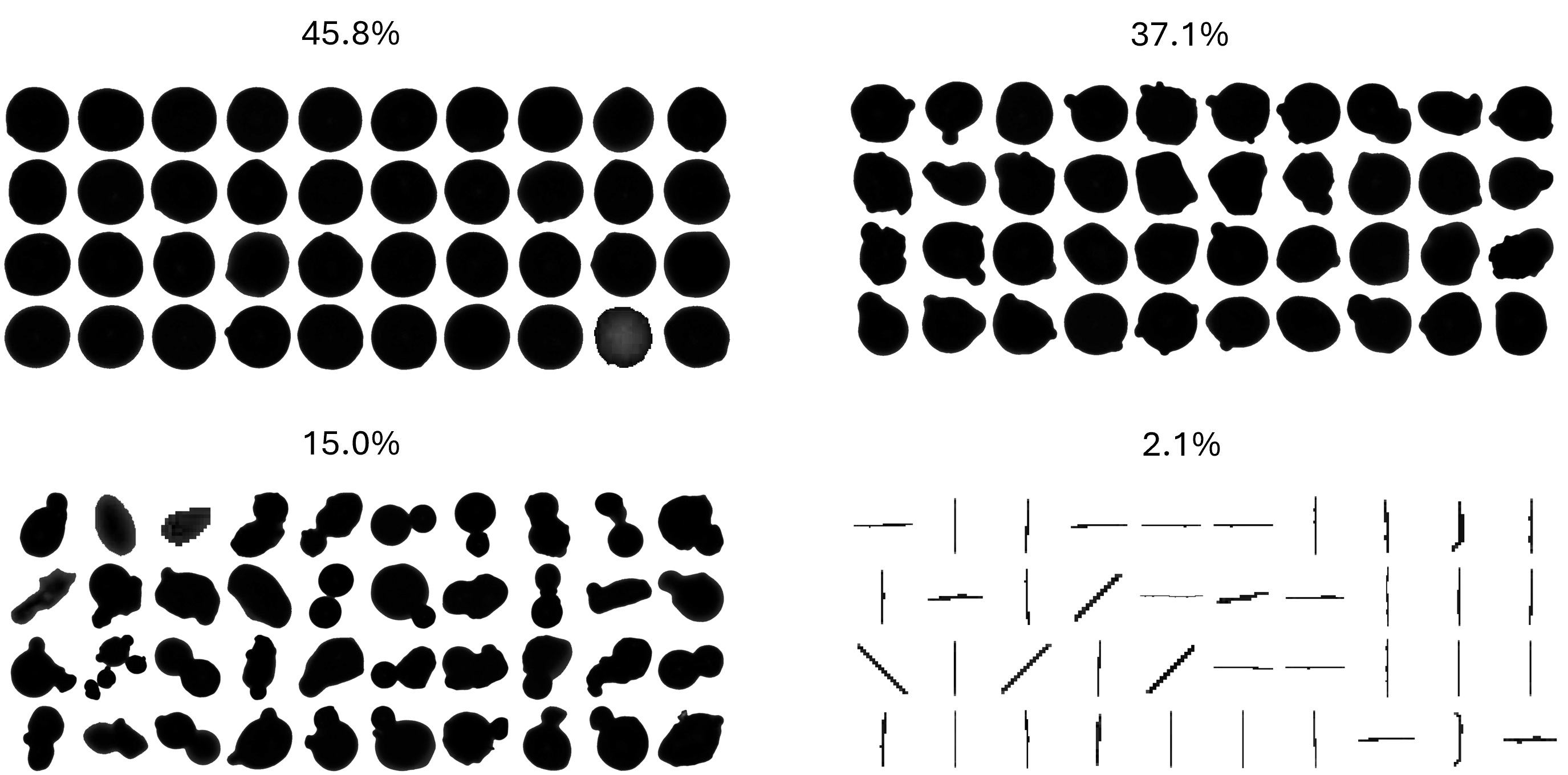}
    \caption{Representative particle shapes for each GMM cluster (based on FD features), with percentages showing each cluster’s share of the dataset.}
    \label{fig:fd-gmms}
    \end{subfigure}
    \hfill
    \begin{subfigure}[c]{0.48\linewidth}
    \centering
    \includegraphics[width=\linewidth]{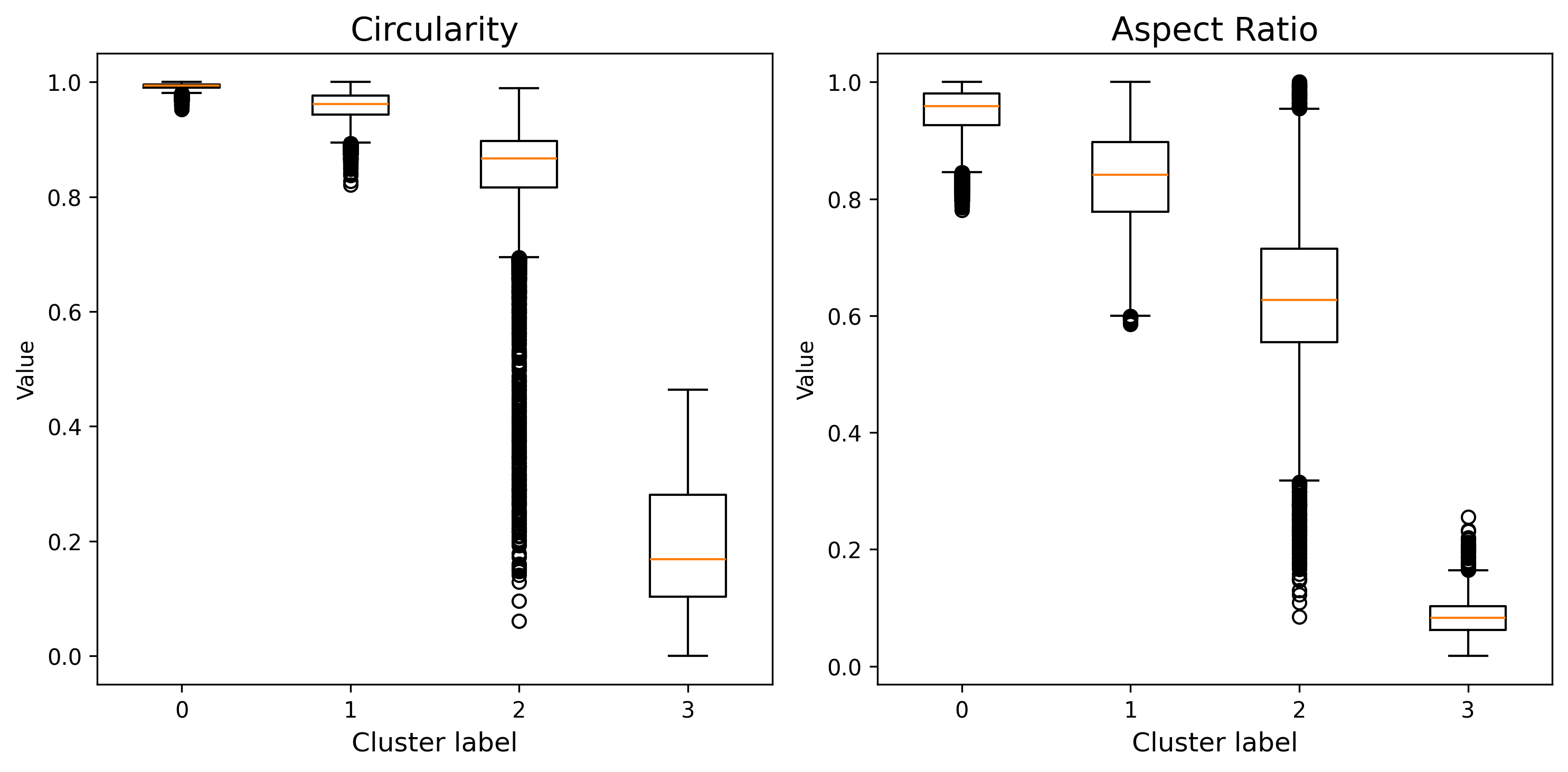}
    \caption{Boxplots of circularity and aspect ratio across the GMM clusters.}
    \label{fig:fd_gmm_boxplot}
    \end{subfigure}
    
    \caption{Both clustering methods identify four clusters in the FD-feature data.}
    \label{fig:fd-samples}
\end{figure}

\paragraph{Morphology analysis via FD features} The two clustering algorithms are directly applied to the extracted 10-dimensional FD data. Again, k-means produces well-delineated clusters in the Fourier descriptor space, with a DB index of 0.7253 compared to GMM's 0.7988. Moreover, k-means achieves a CH score of 270,252, substantially higher than GMM's 174,367, indicating that its clusters maximize between-group variance while minimizing within-group spread. By contrast, the GMM projection (Figure \ref{fig:fd-labels}, right panel) displays a more diffuse point cloud and overlapping contours, especially along the first principal component. This broader dispersion reflects GMM's ability to fit clusters with distinct covariance structures, but it also introduces uncertainty at the boundaries when the underlying shape classes intermingle in the low-dimensional embedding.

Representative particle masks (Figures \ref{fig:fd-kmeans} and \ref{fig:fd-gmms}) further illustrate these differences. Under k-means, nearly two-thirds of the dataset (63.8\%) comprise highly symmetric, near-spherical particles and satellites -- shapes dominated by low-frequency Fourier components -- while the remaining clusters capture a gradient of deformation: fused aggregates (9.7\%), elongated fragments (2.2\%), and a heterogeneous mix of rounded but irregular agglomerates (24.3\%). By contrast, the first cluster of GMM is almost exclusively perfect disks, the second shows only those with very gentle ripples, the third contains purely multi-lobed, blob-like fragments, and the fourth isolates the slender rods and chains. The nuanced separation achieved by GMM highlights its capacity to detect latent substructures within the shape space, even at the expense of cluster compactness and boundary clarity.

\begin{figure}[!h]
    \begin{subfigure}[c]{\linewidth}
    \centering
    \includegraphics[width=0.6\linewidth]{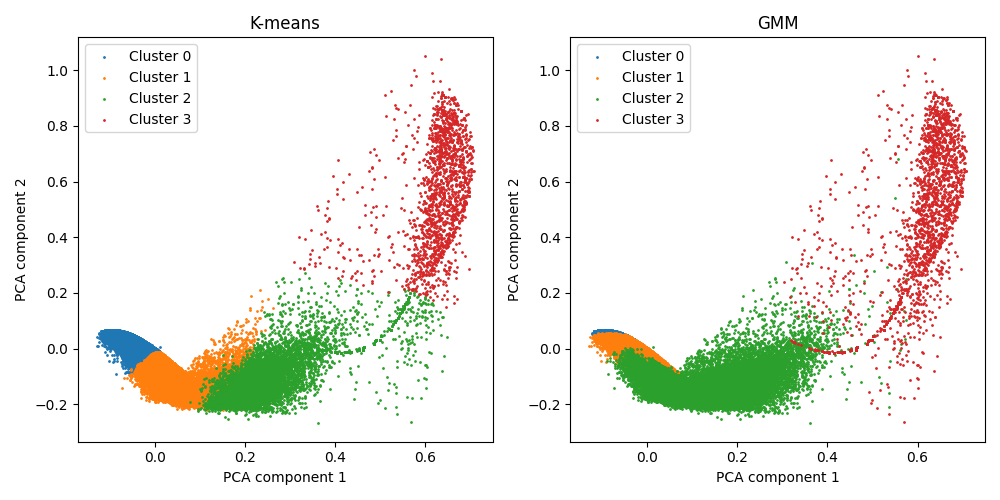}
    \caption{Two-dimensional PCA projection of the 12-dimensional ZM features colored by four k-means clusters (left) and GMM clusters (right).}
    \label{fig:zm-labels}
    \end{subfigure}
    
    \begin{subfigure}[c]{0.48\linewidth}
    \includegraphics[width=\linewidth]{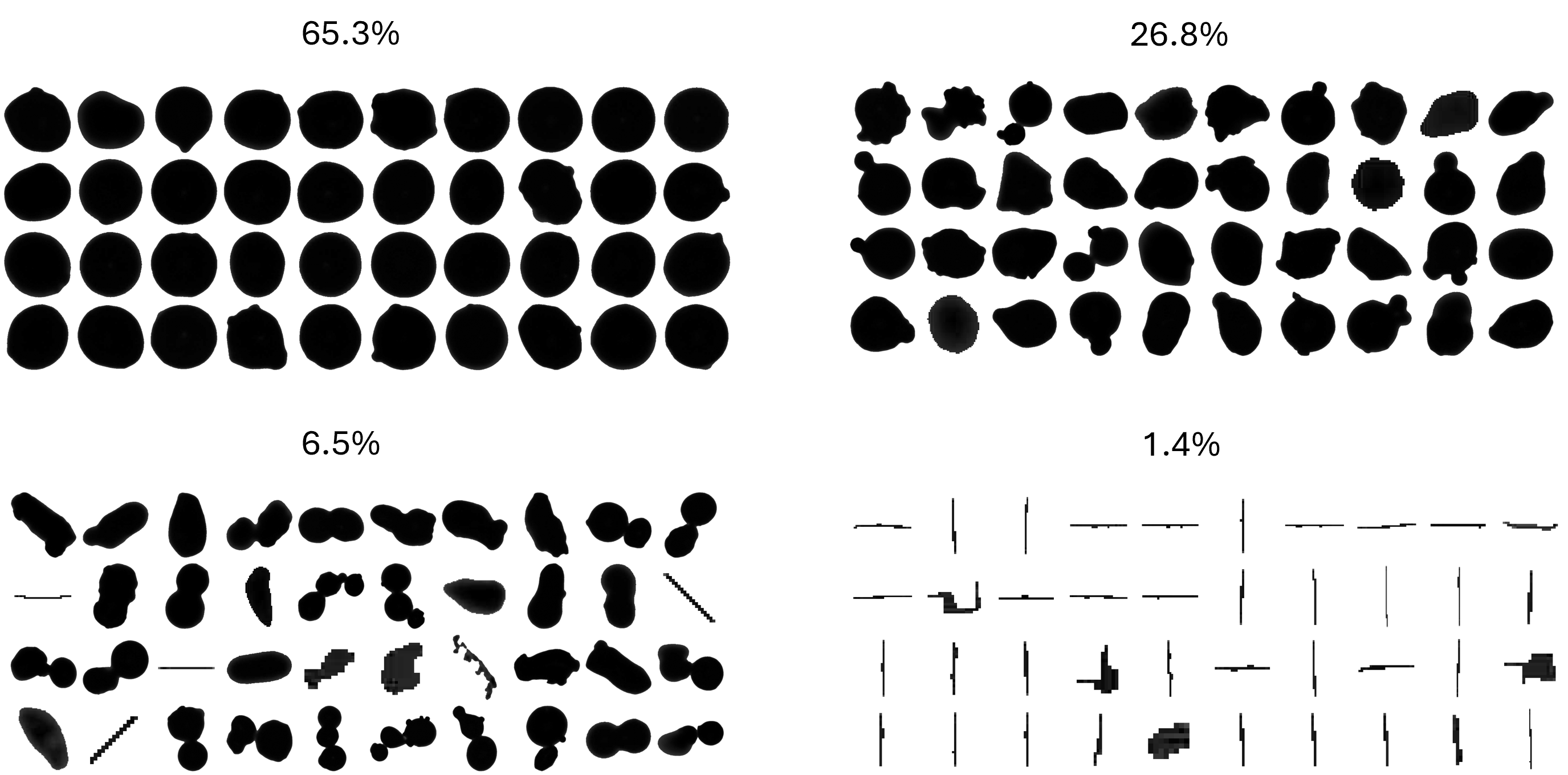}
    \caption{Representative particle shapes for each k-means cluster (based on ZM features), with percentages showing each cluster’s share of the dataset.}
    \label{fig:zm-kmeans}
    \end{subfigure}
    \hfill
    \begin{subfigure}[c]{0.48\linewidth}
    \centering
    \includegraphics[width=\linewidth]{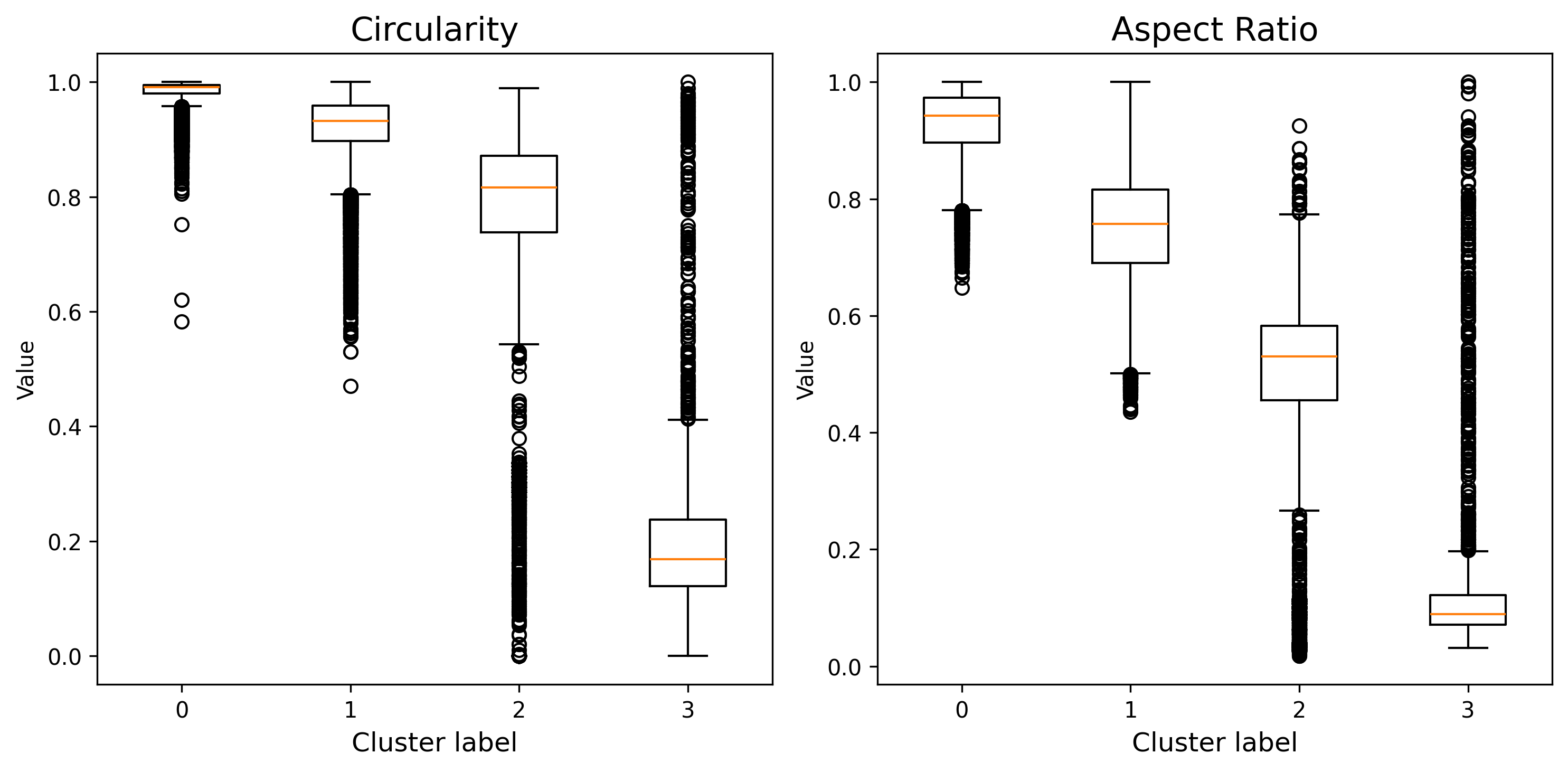}
    \caption{Boxplots of circularity (higher = more round) and aspect ratio (width/length) across the k-means clusters.}
    \label{fig:zN_kmeans_boxplot}
    \end{subfigure}
    
    \begin{subfigure}[c]{0.48\linewidth}
    \includegraphics[width=\linewidth]{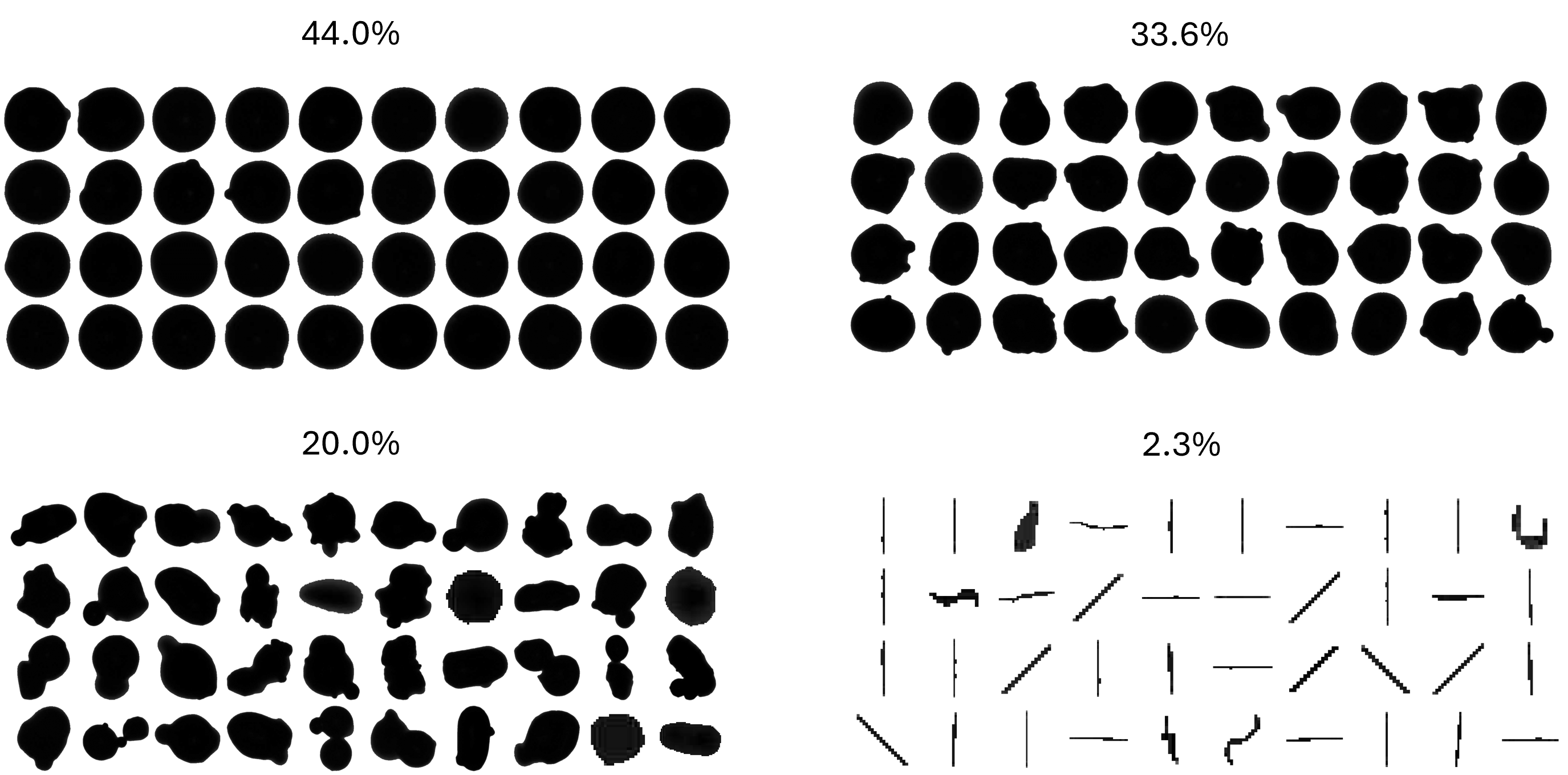}
    \caption{Representative particle shapes for each GMM cluster (based on ZM features), with percentages showing each cluster’s share of the dataset.}
    \label{fig:zm-gmms}
    \end{subfigure}
    \hfill
    \begin{subfigure}[c]{0.48\linewidth}
    \centering
    \includegraphics[width=\linewidth]{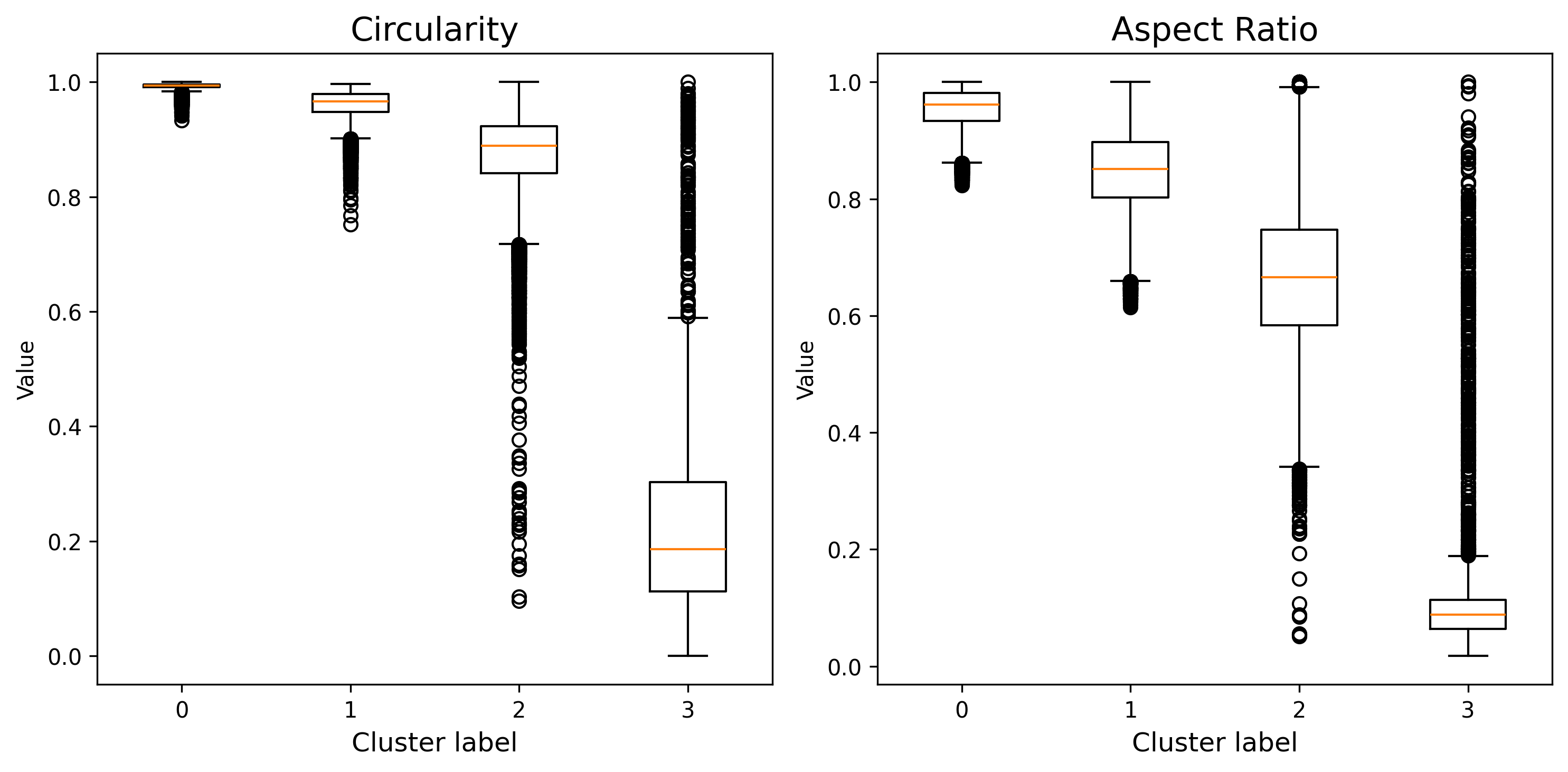}
    \caption{Boxplots of circularity and aspect ratio across the GMM clusters.}
    \label{fig:zm_gmm_boxplot}
    \end{subfigure}
    
    \caption{Both clustering methods identify four clusters in the ZM-feature data.}
    \label{fig:zm-samples}
\end{figure}

\paragraph{Morphology analysis via ZM features}
For each particle image, Zernike moments up to order $n=5$ are computed, producing a 12-dimensional rotation- and scale-invariant shape descriptor. Clustering is then performed directly in this Zernike-moment space without any additional dimensionality reduction. Using this representation, k-means achieves a DB index of 0.8659 and a CH score of 127,426, while GMM yields a DB index of 1.2520 and a CH score of 43,091.
The left plot in Figure \ref{fig:zm-labels} shows the 2-dimensional PCA projection of k-means clustering results, where clusters are compact and separated along linear boundaries, reflecting k-means' tendency to form spherical groups. Cluster 0 and 1 are well confined in the lower left, while Cluster 2 stretches through the middle, overlapping with both Cluster 1 and Cluster 3. Cluster 3 occupies the upper-right region and is relatively distinct. In contrast, the right plot shows the GMM clustering in the same PCA space, where clusters adapt to more elongated, elliptical shapes. GMM places more points into Cluster 2 across the central region, indicating a more fluid partitioning of the data. Although the DB and CH indices favor the more compact k-means solution, the GMM clusters appear to better follow the gradual density transitions in the Zernike descriptor space, illustrating a trade-off between internal validity scores and a smoother, more flexible segmentation.

Examining the exemplar masks (Figures \ref{fig:zm-kmeans} and \ref{fig:zm-gmms}) further highlights these distinctions.  For k-means clustering, in the major cluster (top-left), powders are almost perfectly circular or gently ovoid, with smooth, compact outlines and minimal surface irregularity. The top‐right cluster shows moderately irregular spheres: shapes remain broadly round but exhibit small bumps, indentations, or slight lobes around their perimeters. In the bottom‐left cluster, forms become distinctly non‐spherical and often multi‐lobed or paired, ranging from short ``dumbbell'' and peanut shapes to broader, blockier clusters of two or three fused beads. Finally, the bottom‐right cluster contains the most elongated and filamentary particles -- thin, rod‐like or needle‐shaped fragments, sometimes with jagged or pixelated edges, that contrast sharply with the bulky shapes above. For GMM clustering, the first cluster consists of nearly perfect, smooth circular discs of uniform size; the next shows similarly round shapes that have developed small bulges and indentations, giving each a slightly blobby outline; the third cluster is composed of wildly irregular, amorphous forms with lobed, elongated contours that lack any clear symmetry; and finally, the last set is dominated by slender, rod‐ or line‐like fragments, appearing as straight or gently curved sticks arrayed in various orientations.

Figures \ref{fig:fd-samples} and \ref{fig:zm-samples} confirm that FD and ZM features encode comparable morphological information. Both yield four main clusters under either clustering method, with similar shape distributions. However, GMM tends to isolate the most perfect spheres into a smaller, purer first cluster, while k-means allows for mild deformations within the first cluster. Furthermore, GMM's fourth cluster is dominated by rod-like fragments, whereas k-means includes a wider mix, including multi-lobed agglomerates. This distinction is also evident in the box plots (e.g., Figures \ref{fig:fd_kmeans_boxplot} and \ref{fig:fd_gmm_boxplot}), where k-means clusters show longer whiskers and more outliers, reflecting greater within-cluster variability.

\subsection{Functional Data Clustering} \label{fd_clustering}
The centroid distance function \(D(\theta)\) (see Figure \ref{fig:cdf}) provides a rotation-invariant, scale-normalizable, and translation-invariant representation of shape. We discretize this continuous function by sampling \(D(\theta)\) at 200 equally spaced angles $\theta_k = 2\pi k/200,\ k = 0,\ldots,199,$ where the starting angle is aligned with the longest radius, i.e., \(D(0^\circ)\)=\(\max\{D(\theta_k): \ k = 0,\ldots,199\}\). To convert these discrete samples into smooth functional data, we fit B-spline basis functions to the sequence \(\{D(\theta_0), D(\theta_1), \ldots, D(\theta_{199})\}\), resulting in an open-function representation for each particle.

\begin{figure}[!h]
    \centering
    \includegraphics[width=0.7\linewidth]{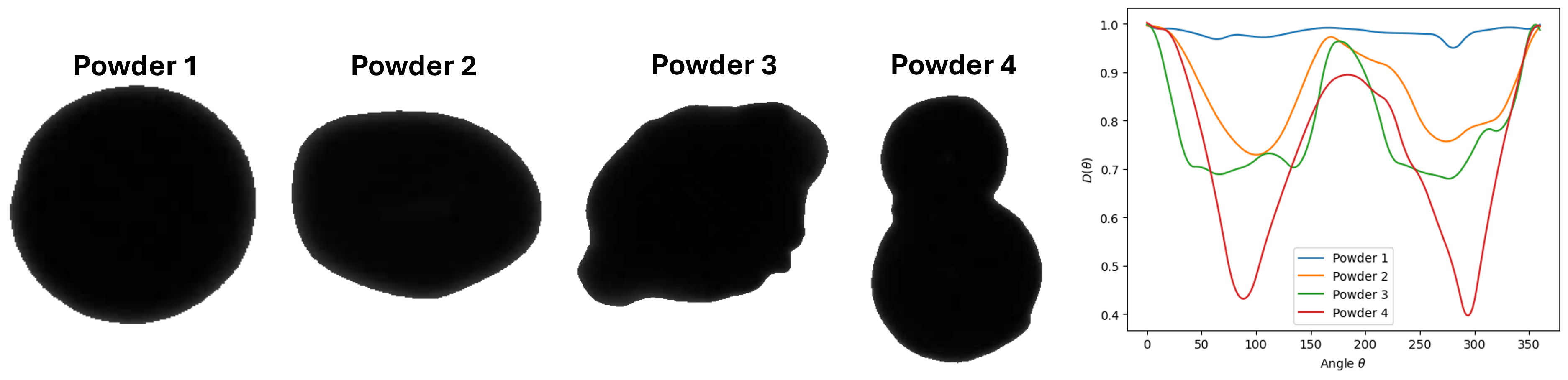}
    \caption{Illustration of functional representation of powder morphology.}
    \label{fig:cdf_examples}
\end{figure}

The centroid distance function $D(\theta)$ provides a direct geometric description of particle shape: for a perfect sphere, $D(\theta)$ is nearly flat, whereas ovoid particles yield gentle undulations, and elongated or fused particles produce deeper troughs at the angles corresponding to indentations or lobes. These variations in the radial profile correspond physically to differences in eccentricity, irregularity, and overall boundary complexity. Although centroid distance functions are well-established shape descriptors, our contribution lies in how they are modeled. Rather than viewing the sampled values of $D(\theta)$ as a high-dimensional vector for use in conventional multivariate clustering (e.g., k-means or GMM), we model $D(\theta)$ as a smooth, continuous function defined on the angle domain $[0, 2\pi)$. This functional perspective preserves the inherent continuity, periodicity, and correlation structure of the radial signal, allowing GPmix to operate directly in the function space where these shapes naturally reside. In contrast, modeling $D(\theta)$ as a simple vector implicitly assumes that consecutive (discrete) samples are independent coordinates, ignoring the fact that neighboring angles encode coupled geometric changes in the outline. The functional representation retains these local dependencies, capturing both the shape and its directional rate of change, which are essential for distinguishing subtle variations and small-scale irregularities in powder morphology. By leveraging this functional representation, GPmix can identify structure and variation in the radial profile that are often attenuated or lost when the same data are treated as an unconstrained vector. Example functions illustrating these patterns are shown in Figure \ref{fig:cdf_examples}.

Given the large dataset size (about 126,000 functions), applying GPmix directly to the full functional dataset is not limited by runtime but by memory. GPmix requires constructing a full pairwise affinity matrix, which stores the similarity between every pair of functions. For a dataset of size $N$, this matrix contains $N^2$ entries; for our dataset, this corresponds to more than $10^{10}$ pairwise similarities, far exceeding the memory capacity of a standard workstation. To address this, we adopted the hybrid sampling strategy introduced in \cite{huang2019ultra}, which combines random sampling with k-means--based prototype selection. In this approach, an initial subset of $N_r$ functions is drawn uniformly at random to capture broad variability, and these are then clustered into $N_k$ groups using k-means; the resulting $N_k$ centroids serve as exemplar functions that summarize the dominant morphological patterns in the full dataset. In our case, we set $N_r$ and $N_k$ to approximately 30\% and 5\% of the dataset, respectively, yielding about 6,300 exemplar functions that preserve the global morphological variability. GPmix is run only on this exemplar subset, after which a functional classifier is trained to infer cluster labels to the full dataset. This two-stage strategy enables GPmix to scale to industrial-sized datasets while avoiding the quadratic memory bottleneck.

GPmix was configured with 12 projection functions generated from the Ornstein–Uhlenbeck process, which provides smooth stochastic basis functions well-suited for functional representations of particle outlines. Using its own BIC mechanism for estimating number of clusters, GPmix identified five clusters within the exemplar set. After obtaining these labels, we trained a functional nearest-centroid classifier on these labeled exemplar functions and used it to assign cluster labels to the remaining samples. This extends the clustering solution to the full dataset without constructing the full $N^2$ affinity matrix required by GPmix.

\begin{figure}[!h]
    \begin{subfigure}[c]{\linewidth}
    \centering
    \includegraphics[width=0.4\linewidth]{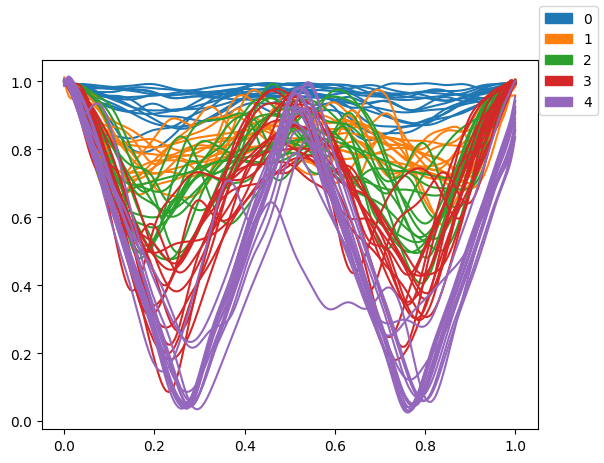}
    \caption{Randomly selected normalized centroid‐distance functions from different clusters. Functions progress from near-flat (Cluster 0) to increasingly undulated profiles (Cluster 4).}
    \label{fig:gpmix-cluster-functions}
    \end{subfigure}
    
    \begin{subfigure}[c]{0.48\linewidth}
    \includegraphics[width=\linewidth]{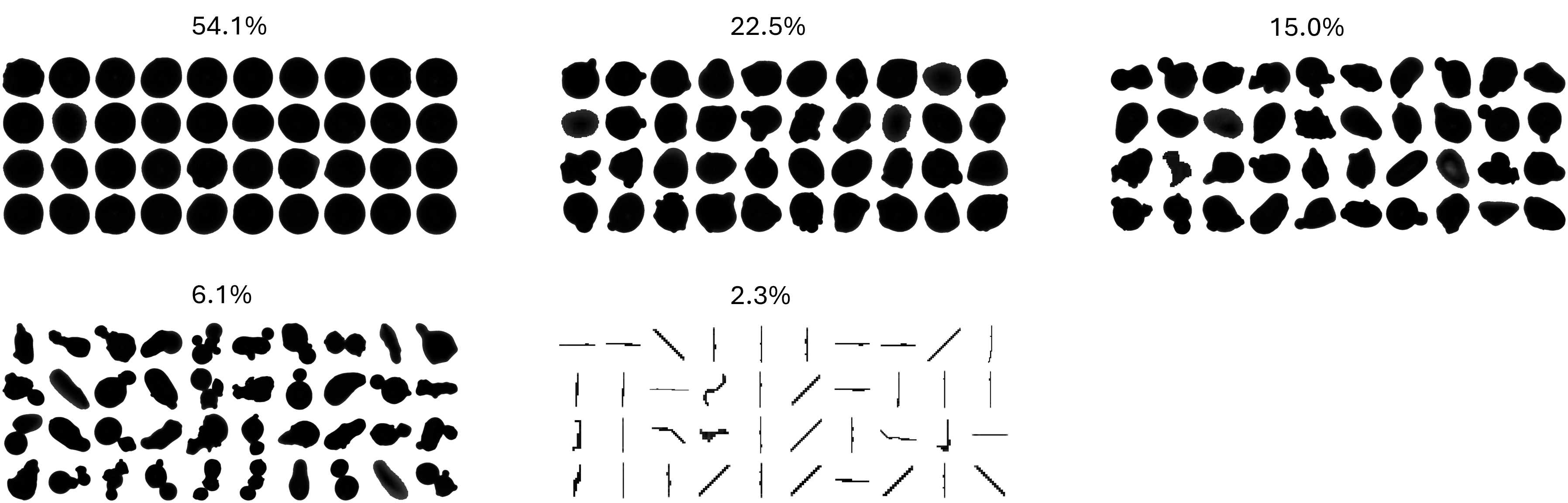}
    \caption{Representative particle shapes for each GPmix cluster, with percentages showing each cluster’s share of the dataset.}
    \label{fig:gpmix}
    \end{subfigure}
    \hfill
    \begin{subfigure}[c]{0.48\linewidth}
    \centering
    \includegraphics[width=\linewidth]{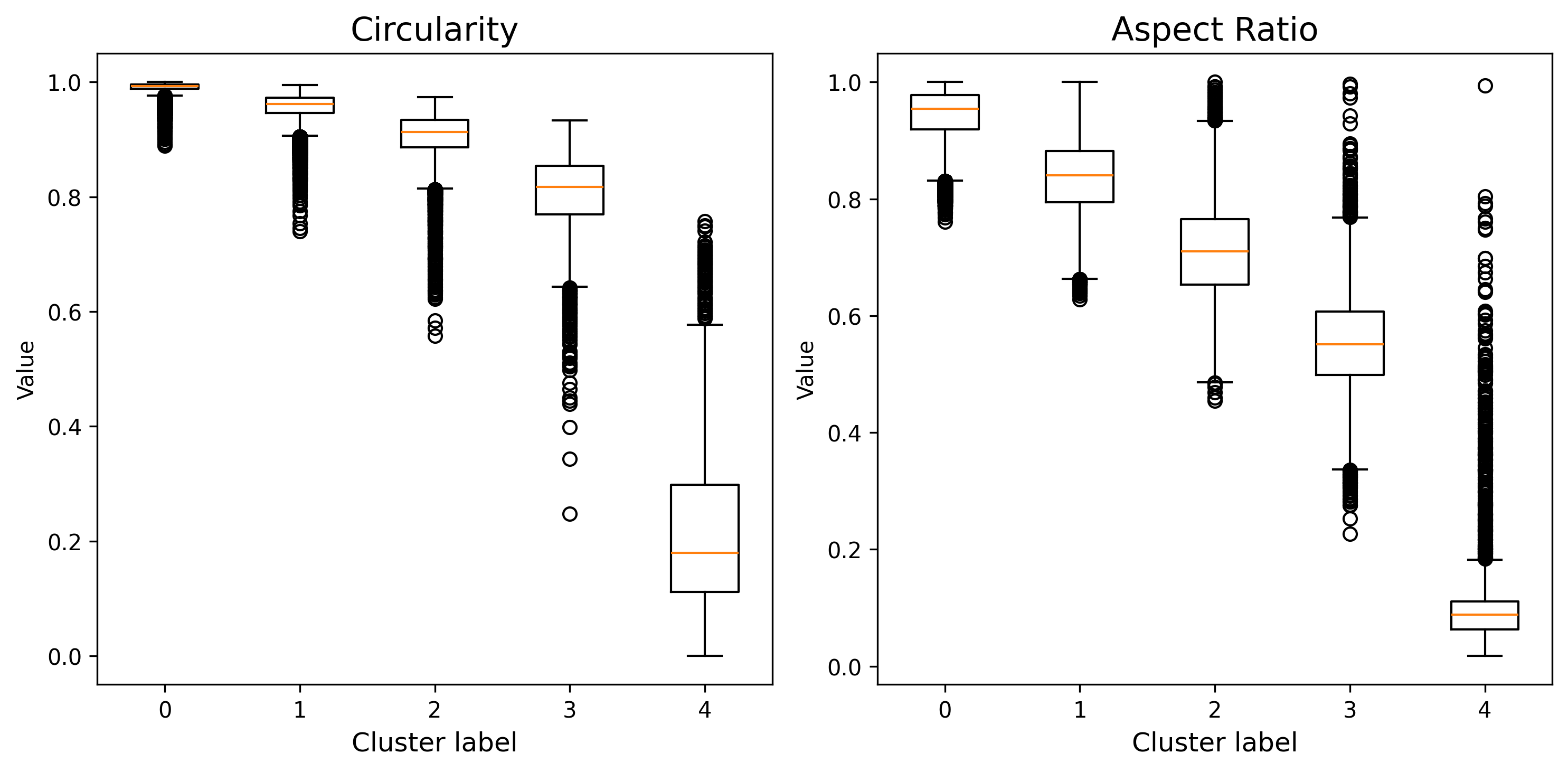}
    \caption{Boxplots of circularity and aspect ratio for each cluster, illustrating decreasing circularity and aspect ratio from Cluster 0 through Cluster 4.}
    \label{fig:gpmix_boxplot}
    \end{subfigure}
    
    \caption{GPmix identifies five clusters in the functional data.}
    \label{fig:gpmix-samples}
\end{figure}

Cluster assignments were evaluated using internal validity indices: the DB index (1.0421) and the CH score (151,832). These scores fall between those obtained by k-means and GMM clustering on the CDF feature dataset. Randomly selected functions from the five clusters are visualized in Figure \ref{fig:gpmix-cluster-functions}. Cluster 0 (blue) contains nearly flat curves, indicating highly spherical particles with uniform radial profiles. Cluster 1  (orange) shows shallow W-shaped curves, corresponding to mildly ovoid or slightly irregular shapes. Cluster 2 (green) has more pronounced W-patterns, suggesting moderate deformation or mild agglomeration. Cluster 3 (red) exhibits sharper dips, indicative of elongated or fused particles with lobed contours. Cluster 4  (purple) displays the deepest and most jagged W-shapes, representing highly elongated, fragmented, or needle-like particles. Representative particles for each cluster, shown in Figure \ref{fig:gpmix}, further illustrate the morphological differences among the identified clusters. The boxplots reveal that, while circularity varies modestly across clusters and aspect ratio show more pronounced differences, underscoring their importance in differentiating shape classes. Notably, the last two GPmix clusters exhibit similar shape distributions to the final two clusters obtained via k-means clustering on the VAE embeddings, suggesting convergence in the identification of extreme morphological subtypes.

\subsection{Computational Efficiency Evaluation}
To compare the computational efficiency of the different pipelines for large-scale morphology profiling, we benchmarked the end-to-end runtime of each method under a consistent experimental setup. For fairness, the number of clusters was fixed at $K=4$ across all methods, regardless of the optimal value identified earlier for any individual pipeline. Each experiment was repeated five times, and the time required for feature extraction and clustering was recorded for every run. Table~\ref{tab:compute_time} reports the mean runtime, with standard deviation shown in parentheses. Feature extraction and clustering were timed separately to allow direct comparison across pipelines with different computational characteristics.

All computations were performed on a standard desktop workstation equipped with an Intel Core i9-12900K CPU (16 cores, 32 GB RAM) and an NVIDIA RTX A2000 (12 GB) GPU. With the exception of VAE training and inference, all computations were executed on the CPU.

\begin{table}[!ht]
\centering
\caption{Mean runtime (seconds) for feature extraction and clustering across pipelines. 
Standard deviation is shown in parentheses. All experiments repeated five times.}
\label{tab:compute_time}

\begin{tabular}{lcccc}
\hline
Feature     & Feature Extraction  & K-means      & GMM          & GPmix         \\
\hline
VAE              & 858.13 (5.17)  & 0.68 (0.04)  & 46.32 (5.88) &  --           \\
FD               & 19.74 (1.44)   & 0.14 (0.01)  & 17.31 (0.86) &  --           \\
CDF-100          & 19.27 (0.13)   & 0.67 (2.10)  & 18.72 (0.37) &  --           \\
ZM               & 80.80 (2.58)   & 0.18 (0.01)  & 26.05 (2.73) &  --           \\
CDF-200          & 19.95 (0.16)   & --           &  --          & 83.90 (2.39)  \\
\hline
\end{tabular}
\end{table}

Among all pipelines evaluated, the Fourier descriptor + k-means combination achieves the lowest total runtime, confirming its suitability for large-scale morphology profiling under practical computational constraints. The Fourier descriptor + k-means pipeline processes particles at approximately 0.16 ms/particle (over 6,000 particles per second), well below the 1 ms/particle range. Training the VAE required approximately 72 hours; therefore, only inference-time generation of latent features is reported in Table~\ref{tab:compute_time}. Although VAE inference is substantially slower than the classical descriptors, the model provides the additional benefit of generative capability, enabling the synthesis of realistic powder shapes that may be useful for certain powder-metallurgy applications. The GPmix pipeline exhibits the largest clustering time because its runtime includes transforming the discrete 200-dimensional centroid distance function (CDF-200) into smooth functional representations, as well as the hybrid sampling and functional classification steps described in Section~\ref{fd_clustering}. Finally, the scalability of k-means is evident across all descriptor families, reinforcing its suitability for high-throughput industrial workflows.

\section{Conclusion}\label{Conclusion}
We have presented and evaluated three complementary, unsupervised pipelines for large-scale morphology profiling of SLM feedstock: (1) an O(2)-invariant variational autoencoder that embeds particle silhouettes into a rotation- and reflection-invariant latent space; (2) a descriptor-based approach that extracts translation-, rotation-, and scale-invariant Fourier, Zernike, and centroid-distance features, followed by conventional clustering; and (3) a functional-data framework (GPmix) that directly clusters radial distance functions. All three methods reliably distinguished key morphological classes: highly spherical particles, those with small satellites or surface indentations; highly irregular, lobed, or elongated shapes; and slender, rod-like fragments. Compared to k-means clustering, the GMM tends to more accurately distinguish both perfectly spherical particles and slender, rod-like fragments which may represent actual powder defects or artefacts introduced by the imaging or analysis system.

While the steerable VAE delivered a flexible, orientation-invariant representation, its high computational and memory demands make it impractical for real-time quality monitoring. The GPmix pipeline preserved fine-grained shape details and achieved strong clustering validity, but its quadratic scaling memory requirement and two-stage exemplar workflow hinder straightforward deployment at scale. In contrast, the Fourier-descriptor + k-means combination achieved excellent separation of spherical, lobed, and elongated classes with minimal computational overhead (\(<\) 1 ms/particle), offering the best balance of accuracy, speed, and implementation simplicity.

Accordingly, we recommend the Fourier descriptor + k-means pipeline as the primary tool for continuous, high-throughput powder morphology assessment within SLM environments. The VAE and GPmix methods remain valuable for exploratory analyses or specialized adaptive-inspection scenarios, where their unique representational and clustering capabilities can uncover subtler shape variations beyond routine monitoring needs.

Because many alloy systems used in LPBF are produced via similar atomization or milling routes, their feedstock powders exhibit broadly comparable particle morphologies, making the proposed framework readily transferable to other powder chemistries. The methodology developed here also provides a practical basis for real-time monitoring of powder quality across successive reuse cycles.  Once an initial morphology distribution is obtained at the start of a build, the clustered dataset can be used to train a supervised classifier capable of recognizing these morphology classes in new particle images. During subsequent cycles, in-situ powder images can be processed through the same invariant feature-extraction pipeline (e.g., Fourier descriptors) and passed to the classifier to estimate the evolving proportions of each morphology class. Tracking these temporal shifts provides a quantitative means of assessing powder degradation, identifying the accumulation of defective or irregular particles, and determining when powder refreshment is required. Coupled with mechanical or process-performance data, this monitoring strategy has clear potential for industrial deployment. Future work may extend this framework to three-dimensional imaging modalities, such as X-ray computed tomography or holographic methods, to capture additional geometric detail and to validate morphology distributions inferred from 2D projections.

\section*{Author contributions: CRediT}
\noindent
\textbf{Emmanuel Akeweje}: Methodology, Software, Validation, Formal analysis, Writing - Original Draft, Visualization; \textbf{Conall Kirk}: Data Curation, Writing - Original Draft, Visualization; \textbf{Chi-Wai Chan}: Conceptualization, Data Curation, Writing - Review \& Editing, Supervision; \textbf{Denis Dowling}: Conceptualization, Writing - Review \& Editing, Supervision; \textbf{Mimi Zhang}: Conceptualization, Methodology, Writing - Review \& Editing, Visualization, Supervision.

\section*{Funding sources}
This publication has emanated from research conducted with the financial support of Research Ireland under Grant number 21/RC/10295\_P2 and Queen's University Belfast under Grant number R8448MEE. For the purpose of Open Access, the author has applied a CC BY public copyright license to any Author Accepted Manuscript version arising from this submission. 

\section*{Data availability}
Data will be made available on request.
\bibliographystyle{elsarticle-num-names} 
\bibliography{references}
\end{document}